\documentclass[letterpaper]{article} % DO NOT CHANGE THIS
\usepackage{aaai25}  % DO NOT CHANGE THIS
\usepackage{times}  % DO NOT CHANGE THIS
\usepackage{helvet}  % DO NOT CHANGE THIS
\usepackage{courier}  % DO NOT CHANGE THIS
\usepackage[hyphens]{url}  % DO NOT CHANGE THIS
\usepackage{graphicx} % DO NOT CHANGE THIS
\usepackage{natbib}  % DO NOT CHANGE THIS AND DO NOT ADD ANY OPTIONS TO IT
\usepackage{caption} % DO NOT CHANGE THIS AND DO NOT ADD ANY OPTIONS TO IT
\usepackage{algorithm}
\usepackage{algorithmicx}
\usepackage{algpseudocode}
\usepackage{newfloat}
\usepackage{listings}
\DeclareCaptionStyle{ruled}{labelfont=normalfont,labelsep=colon,strut=off} % DO NOT CHANGE THIS
\floatstyle{ruled}
\newfloat{listing}{tb}{lst}{}
\floatname{listing}{Listing}
\title{AAAI Press Formatting Instructions \\for Authors Using \LaTeX{} --- A Guide}
\title{My Publication Title --- Multiple Authors}
\author {
        Mateusz Gajewski \equalcontrib \textsuperscript{\rm 2, \rm 3}, Adam Karczmarz \equalcontrib \textsuperscript{\rm 1, \rm 2}, Mateusz Rapicki \equalcontrib \textsuperscript{\rm 1}, Piotr Sankowski\textsuperscript{\rm 1, \rm 4}
}
\affiliations {
    \textsuperscript{\rm 1}  Faculty of Mathematics, Informatics and Mechanics University of Warsaw, Warsaw, Poland\\
    \textsuperscript{\rm 2}IDEAS NCBR\\
    \textsuperscript{\rm 3} Faculty of Computing and Telecommunications Poznan University of Technology, Poznan, Poland\\
    \textsuperscript{\rm 4} MIM Solutions 

    mg96272@gmail.com, a.karczmarz@mimuw.edu.pl, mateusz.rapicki@mimuw.edu.pl, sank@mimuw.edu.pl
}

\usepackage{bibentry}
\usepackage{graphicx}
\usepackage{amssymb,amsmath,amsfonts,amsthm}
\usepackage{todonotes}
\usepackage{thmtools,thm-restate}
\usepackage{booktabs}
\usepackage{enumitem}

\title{Accurate Estimation of Feature Importance Faithfulness for Tree Models}
\date{}

\begin{document}

\newcommand{\adam}[1]{{\color{blue}{\textsc{Adam}: #1}}}
\newcommand{\mg}[1]{{\color{orange}{\textsc{MateuszG}: #1}}}
\newcommand{\rapicki}[1]{{\color{olive}{\textsc{Rapicki}: #1}}}

\maketitle

\newcommand{\tr}{\mathcal{T}}
\newcommand{\tf}{q}
\newcommand{\lvs}{\mathcal{L}}
\newcommand{\perturb}{\text{perturb}}
\newcommand{\PGI}{\mathrm{PGI}}
\newcommand{\PG}{\mathrm{PG}}
\newcommand{\EX}{\mathbb{E}}
\newcommand{\dist}{\mathcal{D}}
\newtheorem{theorem}{Theorem}[section]

\begin{abstract}

  In this paper, we consider a perturbation-based
  metric of predictive faithfulness of feature rankings (or attributions)
  that we call \emph{PGI squared}. 
  When applied to decision tree-based regression models, the metric can be computed \emph{exactly}
  and~efficiently
  for arbitrary independent feature perturbation distributions. In particular,
  the computation does not involve Monte Carlo
  sampling
  that has been typically used for computing similar metrics and which is
  inherently prone to inaccuracies.

As a second contribution, we proposed a procedure for constructing feature ranking based on PGI squared. Our results %of our comparative analysis
  indicate the proposed ranking method is comparable to the widely recognized SHAP explainer, offering a viable alternative for assessing feature importance in tree-based models. %\adam{mozna sie jeszcze zastanowic, czy taka jest konkluzja}
\end{abstract}

\begin{links}
     \link{Code}{https://github.com/rapicki/prediction-gap}
     \link{Extended version}{https://arxiv.org/abs/2404.03426}
\end{links}
\section{Introduction}\label{sec:intro}
One of the key challenges in deploying modern machine learning models in such areas as medical diagnosis lies in the ability to indicate why a certain prediction has been made. Such an indication may be of critical importance when a human decides whether the prediction can be relied on. 
This is one of the reasons various aspects of explainability of machine learning models have been the subject of extensive research lately (see, e.g.,~\citep{BurkartH21}).

For some basic types of models (e.g., single decision trees,
the rationale behind a prediction is easy to understand by a human. However, predictions of more complex models (that offer much better accuracy, e.g., based on neural networks or decision tree ensembles) are also
much more difficult to interpret. Accurate and concise explanations understandable to humans might not always exist. 
In such cases,
it is still beneficial to have methods
\emph{giving a flavor} of what factors might have influenced the prediction the most.

Local feature attribution\footnote{It should be contrasted with \emph{global} feature attribution whose aim is to measure the feature's total impact on the model.} constitutes one of such general approaches. For a fixed input $x$ and the model's prediction $f(x)$, each feature is assigned a weight ``measuring'' the feature's impact on the prediction. 
Ideally, a larger (absolute) weight should correspond to a larger importance of the feature for the prediction. 

Many local attribution methods have been proposed so far, e.g., \citep{SHAP, PlumbMT18, Ribeiro0G16}. They are often completely model-agnostic, that is, they can be defined for and used with any model which is only accessed in a black-box way for computing the attribution. In some of these cases, focusing on a particular model's architecture can lead to much more efficient and accurate attribution computation algorithms (e.g.,~\citep{LundbergECDPNKH20} for tree ensemble models).

With the large body of different attribution methods, it is not clear which one should one use, especially since there is not a single objective measure of their reliability. To help deal with this issue, several notions of explanation quality have been proposed in the literature, such as faithfulness (fidelity)~\citep{LiuKWN21, yeh2019fidelity}, stability~\citep{alvarez2018robustness}, or fairness~\citep{balagopalan2022road, dai2022fairness}. 
\citeauthor{agarwal2022openxai} \shortcite{agarwal2022openxai} developed an open-source benchmark OpenXAI automatically
computing variants of these measures for a number of proposed attribution methods.
 
Out of these notions, our focus in this paper is on faithfulness.
Roughly speaking, an attribution method deserves to be called faithful if the features deemed important by the attribution are truly important for the decision-making process of the model. The \emph{perturbation-based} methods constitute one popular class of approaches to measuring faithfulness. Intuitively, perturbing features deemed impactful should generally lead to a significant change in the prediction. On the contrary, manipulating unimportant features should not make a big difference.

One concrete cleanly-defined perturbation-based faithfulness metric for regression problems is the \emph{prediction gap on important feature perturbation}, PGI in short. PGI is a faithfulness measure of choice in the OpenXAI benchmark~\citep{agarwal2022openxai}. Having fixed some subset of important features $S$ derived from the obtained attribution, the \emph{prediction gap} (PG) of $x$ wrt. $S$ is defined as:
\begin{equation}\label{eq:pg}
\PG(x,S):=\EX_{x'\sim \perturb(x,S)}\left[|f(x')-f(x)|\right].
\end{equation}
In~\citep{agarwal2022openxai}, the concrete perturbation method $\perturb(x,S)$ is adding 
an independent Gaussian noise from $\mathcal{N}(0,\sigma^2)$ to each coordinate $x_j$, $j\in S$.
$\PGI(x)$ can be defined as either the prediction gap wrt. a fixed set of important features (e.g., $k$ top-scoring features for some fixed~$k$~\citep{dai2022fairness}), or an average prediction gap over many important feature sets (as in~\cite{agarwal2022openxai}).
In particular, the PGI metric depends \emph{only} on the ordering of features by importance induced by the attribution.
Other perturbation-based methods include e.g., PGU~\citep{agarwal2022openxai}, comprehensiveness, sufficiency~\citep{DeYoungJRLXSW20}, remove-and-retrain~\citep{HookerEKK19}, deletion~\citep{PetsiukDS18}.

In practice, PGI and other methods based on random perturbations are computed using the Monte Carlo method which is inherently prone to inaccuracies, unless a large number of samples is used. 

It is interesting to ask whether PGI or similar random perturbation-based faithfulness metrics can be computed \emph{exactly}, that is, via a closed-form calculation, so that any approximation error is caused merely by the floating-point arithmetic's rounding errors.

\subsection{Our Contribution}
\subsubsection{Exact Computation} First, we consider random perturbation-based measurement of faithfulness from the point of view of
efficient and accurate computation. Of course, it is unrealistic to assume that such a quantity
can be computed exactly (or even beyond the Monte Carlo method) given only black-box access to the model
and the perturbation distribution. Hence, we focus on a concrete model architecture:
\emph{tree ensemble models}. Tree ensemble models remain a popular choice among practitioners
since they are robust, easy to tune, and fast to train.

Unfortunately, computing $\PG(x,S)$ (wrt. a fixed set $S$ of perturbed features, as in~\cite{agarwal2022openxai}, see~\eqref{eq:pg})
does not look very tractable even given the rich structure that tree ensemble models offer. This is because
therein, the expectation is taken over the \emph{absolute value} of $f(x)-f(x')$,
where $x'$ is $x$ with important features perturbed.
The absolute value is not very mathematically convenient to work with when taking expectations.
On the other hand, dropping the absolute value here would not make much sense: a small random
perturbation (positive or negative) of a feature could in principle lead to a large positive or negative prediction difference. These, in turn,  could cancel out in expectation, making
the metric close to zero even though the feature is visibly of high importance.

\vspace{2mm}
\emph{Theoretical contribution:} To deal with the tractability problem of measuring PGI on tree models, we propose a similar metric,
$\PGI^2$, obtained from a slightly different \emph{squared prediction gap}:
\begin{equation}\label{eq:pgsquared}
\PG^2(x,S):=\EX_{x'\sim \perturb(x,S)}\left[(f(x')-f(x))^2\right].
\end{equation}
This eliminates the cancellation problem outlined above.
Of course, introducing the square can alter the evaluation of the feature's importance
wrt. PGI as defined previously. However, intuitively, it amplifies the big prediction differences, and at the same
time marginalizes the small, which is a generally good property if we seek concise explanations. 

We prove that for tree ensemble models with $n$ nodes in total, the squared prediction gap $\PG^2(x,S)$ can be computed
in $O(n^2)$ time \emph{exactly} for any fixed choice of a subset $S$ of perturbed important features, assuming the cumulative distribution function of feature-wise random perturbations can
be evaluated in constant time.
That is, the obtained running time bound does not require that the random perturbations are Gaussian or their distributions are equal across features.

We note that whereas our $\PG^2$ algorithm is quadratic, from the theoretical point of view, the important thing is that on tree models, a prediction gap-style quantity computation can be carried out \emph{exactly} in \emph{polynomial time} after all.

Given a permutation $\pi$ ranking the $d$ features from the most to least important, the OpenXAI benchmark calculates $\PGI$ by taking the average prediction gap
obtained when perturbing $k=1, 2, \ldots, d$ most important features $\pi[1..k]=\{\pi(1),\ldots,\pi(k)\}$ (see~\citep[Appendix~A]{agarwal2022openxai}). In such a case, $d$ different
expectations of absolute prediction gap are estimated, each via Monte Carlo sampling.
In our case, $\PGI^2(x)$ is analogously defined as:
\begin{equation}\label{eq:pgisquared}
\PGI^2(x,\pi):=\frac{1}{d}\sum_{k=1}^d\PG^2(x,\pi[1..k]).
\end{equation}

Using our prediction gap algorithm, $\PGI^2$ is computed exactly (assuming infinite precision arithmetic; in reality, up to the precision error incurred by the usage of floating-point arithmetic) in $O(n^2d)$ time. 

\vspace{2mm}
\emph{Experiments:}
In our experiments, we evaluated different methods for calculating the Prediction Gap. Our study encompassed the exact algorithm as well as two sampling techniques integration techniques: Monte Carlo (MC) and Quasi-Monte Carlo (QMC). With the increase in iteration count, the outputs of MC and QMC visibly converged to the output of our exact algorithm; this confirms good numerical stability of our approach.
%\adam{dodalem to zdanie, co sadzicie?} \mg{Myślę że ok jest}

When allocating an equivalent time budget across all methods, our findings revealed an advantage over MC and QMC, with QMC demonstrating a slight edge. Specifically, the Normalized Mean Absolute Error (NMAE) for MC was 0.13
%\adam{czy nmae jest sens podawać w procentach?} \mg{No chyba masz racje - na wykresach tez nie ma \%, wiec zmienie}
for single models and decreased to 0.01 for bigger models. In comparison, QMC exhibited NMAEs of approximately 0.05 for single models and 0.002 for bigger ones. These results underscore the efficacy of sampling methods in the context of computing the PG, particularly for more sophisticated model structures. 
%The discussion of the QMC methodology employed in this study, are in ref{}, which provides an exploration of its technical aspects and implementation details.
%\adam{ostatnie zdanie bym usunal stad} \mg{Ok, ale rozumiem, że opic QMC zostawiamy w appendixie?}\adam{tak}

%We accompany our theoretical development with an experimental comparison of the results
%obtained using the exact algorithm and Monte Carlo sampling. 

%The experiments confirm that Monte Carlo simulation gives noticeably less accurate results within the allocated time (iteration) budget comparable to the running time of our algorithm.
%Specifically, in our experiments, the normalized mean absolute error (NMAE) is typically between $0.02$ and $0.3$. The error measured this way
%generally decreases when increasing the size of the model which follows from the quadratic complexity of our algorithm. It is also decreasing in
%the magnitude of 
%perturbations used: sampling is generally better at estimating larger quantities, whereas tiny perturbations lead to tiny prediction gaps. 

\subsubsection{$\text{PG}^2$-Based Greedy Feature Ordering.}
We also investigate the possibility of using
our exact $\text{PG}^2$ algorithm as a base
of a feature importance ranking algorithm. We stress that finding a feature ordering 
optimizing the $\text{PGI}^2$ metric is a highly non-trivial task, as the metric depends on the entire ordering of the features. Checking all the features' permutations is infeasible in most cases.

We consider constructing the feature ranking using a \emph{greedy} $\PG^2$ heuristic: the $i$-th most important feature $\lambda$ is chosen so that
it optimizes the squared prediction gap together with the already chosen $i-1$ most important features plus $\lambda$. 
We next compare the greedy $\PG^2$ ranking with the ranking produced using the popular SHAP feature attribution method for tree ensembles~\citep{LundbergECDPNKH20}.
We observe that the most important features identified by the two methods generally differ, and the deviation is clearer if the applied perturbations are smaller.

As far as faithfulness is considered, our experiments confirm
that the greedy $\PG^2$ ranking
yields, on average, better $\PGI^2$
scores than SHAP.
While that such a property holds is non-obvious, it is perhaps not very surprising either.
This is why we also
compare the $\PG^2$-based ranking method with SHAP in terms of two different faithfulness metrics: \emph{feature randomization} and \emph{remove-and-retrain}~\cite{HookerEKK19}.

In the former case and some of our datasets, the $\PG^2$-based rankings (used with Gaussian perturbations for small standard deviations) performed better than SHAP, whereas in the others, the results were comparable. 
Wrt. the latter metric, SHAP achieved better performance.

%
%\begin{enumerate}
%\item[(1)] a different faithfulness metric \adam{TODO}, resembling \adam{TODO},
%\item[(2)] average \emph{conciseness}.
%\end{enumerate}
\iffalse
\adam{czy pisać to conciseness} Specifically, we consider three natural metrics of averaged feature importance
across the test data set, each of them based
solely on the feature ordering and clearly promoting the most important features.
For each of these metrics, we compare the entropy of the obtained averaged importances. Intuitively, a smaller entropy indicates that
only few features are identified as critically important to the model.
Our experiments show that the greedy $\PG^2$ ranking used with perturbations of sufficiently large standard deviation can lead to global importance scores with smaller entropy compared to SHAP.
\fi

\section{Preliminaries}\label{s:prelims}

Let us denote by $f:\mathbb{R}^d\to \mathbb{R}$ the output
function of the considered regression model. We sometimes also use
$f$ to refer to the model itself.
The input $x\in\mathbb{R}^d$ to $f$ is called
a feature vector, and we denote by $x_i$ the value of the $i$-th
feature of~$x$. 
Thus, we identify the set of features with~$[d]$.

\paragraph{Tree Models.}
When talking about decision trees, we assume them to be
binary and based on single-value splits. 
That is, each non-leaf node $v$ of a decision
tree $\tr$ has precisely two children $a_v,b_v$.
It is also assigned a feature $\tf_v\in[d]$
and a threshold value $t_v\in\mathbb{R}$.
Each leaf node $l\in \tr$ is in turn assigned
a value $y_v\in\mathbb{R}$.
We denote by $\lvs(\tr)$ the set of leaves of
the tree $\tr$.

The output $f_\tr(x)$~of the tree $\tr$ is computed by following a root-leaf
path in $\tr$: at a non-leaf node $v\in \tr$, we descend either to the child $a_v$ if
$x_{\tf_v}< t_v$, or to $b_v$ otherwise. When a leaf~$l$ is eventually reached,
its value $y_l$ is returned.
We denote by $n_{\tr}$ the number of nodes in $\tr$.

When considering tree ensemble models $(\tr)_{i=1}^m$, the output $f(x)$ of the model is simply the sum of outputs $f_{\tr_i}(x)$ of its~$m$ individual trees. We generally use $n=\sum_{i=1}^m n_{\tr_i}$ to
refer to the \emph{total size} of the tree ensemble model.

\iffalse
\paragraph{Prediction gap on important feature perturbation.}
Let a \emph{ranking} $\pi\in [d]\to [d]$ be a permutation of features
giving their importance from highest to lowest.
Denote by $\pi[1..\ell]$ the set $\{\pi(1),\ldots,\pi(\ell)\}$,
that is, the $\ell$ top ranking features in $\pi$.
For any $x\in \mathbb{R}^d$ and $S\subseteq [d]$, let $\perturb(x,S)$
be some distribution on feature vectors $x'\in\mathbb{R}^d$ satisfying
$x'_i=x_i$ for all $i\notin S$. \cite{agarwal2022openxai} define:
\begin{equation*}
\PGI(x,\pi,k):=\EX_{x'\sim\perturb(x,\pi[1..k])}\left[|f(x')-f(x)|\right].
\end{equation*}
Based on that, they define an averaged metric:
\begin{equation*}
\PGI(x,\pi):=\frac{1}{d}\sum_{k=1}^d \PGI(x,\pi,k).
\end{equation*}
We analogously define the $\PGI^2$ metric:
\begin{align*}
\PGI^2(x,\pi,k)&:=\EX_{x'\sim\perturb(x,\pi[1..k])}\left[\left(f(x')-f(x)\right)^2\right],\\
\PGI^2(x,\pi)&:=\frac{1}{d}\sum_{k=1}^d \PGI^2(x,\pi,k).
\end{align*}
\fi

\section{Computing $\PG^2$ for Tree Ensemble Models}\label{sec:pgi_algorithm}
Let $S\subseteq [d]$ be a subset of \emph{important features} and $x\in \mathbb{R}^d$. In this section, we show our polynomial exact algorithm computing $\PG^2(x,S)$, as defined in~\eqref{eq:pgsquared}.

We consider distributions $\perturb(x,S)$ such that
${x'\sim\perturb(x,S)}$ satisfies:
\begin{enumerate}
\item for every \emph{perturbed feature} $i\in S$, $x'_i=x_i+\delta_i$, where $\delta_i\sim\dist_i$, and distribution $\dist_i$ is fixed, and the c.d.f. $F_{\dist_i}$
of $\dist_i$ can be evaluated in $O(1)$ time.
\item for every \emph{non-perturbed feature} $i\notin S$, $x'_i=x_i$,
\item all $\delta_j$ ($j\in S$) are independent random variables.
\end{enumerate}

The main goal of this section is to show how to compute
\begin{equation*}
\PG^2(x,S):=\EX_{x'\sim \perturb(x,S)}\left[(f(x')-f(x))^2\right]
\end{equation*}
under the assumptions made in the case when $f$ is
a tree ensemble model $(\tr)_{i=1}^m$. Recall from Equation~\eqref{eq:pgisquared} that the metric $\PGI^2(x,\pi)$ for a
given ranking $\pi$ can be computed by running the algorithm~$d$ times for $d$ different subsets $S$ of the form $\pi[1..k]$.

Put $c:=f(x)$. Let $x'\sim\perturb(x,S)$.
Recall that ${f(x')=\sum_{i=1}^m f_{\tr_i}(x')}$. For every leaf node $v$ of some $\tr_i$, consider a random
indicator variable $X_{v}$ such that $X_v=1$
iff $\tr_i$ evaluates to the value $y_v$
of leaf $v$ given $x'$ as input. In particular, for
any $i=1,\ldots,m$, there is exactly one $v\in\lvs(\tr_i)$ such that $X_v=1$.
Set $\lvs=\bigcup_{i=1}^m\lvs(\tr_i)$.
Then:
\begin{equation*}
f(x')=\sum_{v\in\lvs}X_v\cdot y_v.
\end{equation*}
By linearity of expectation, we obtain:
\begin{align*}
\PG^2(x,S)&=\EX\left[\left(\sum_{v\in\lvs}X_vy_v-c\right)^2\right]\\
&=c^2+\EX\left[\left(\sum_{u\in\lvs}X_uy_u\right)\left(\sum_{v\in\lvs}X_vy_v-2c\right)\right]\\
&=c^2+\sum_{\substack{u,v\in \lvs\\u\neq v}}y_uy_v\Pr[X_u=1\land X_v=1]\\
&\qquad\qquad +\sum_{u\in \lvs}\Pr[X_u=1]\cdot y_u\cdot (y_u-2c).
\end{align*}
The above formula reduces computing $\PG^2(x,S)$
to calculating probabilities of the form
$\Pr[X_u=1\land X_v=1]$ (where $u\neq v$) or of the form $\Pr[X_u=1]$.
We focus on the former task as dealing with the latter is analogous but easier.

For a node $w$, let $Q_w$ be the set of features appearing
on the root-$w$ path (in the tree whose $w$ is the node of). 
Observe that for $w\in\lvs$, $X_w=1$ holds
iff for each feature $q\in Q_w$,
$x'_q$ falls into a certain interval $I_{w,q}$.
Namely, if the root-to-$w$ path consists
of nodes $w_1,\ldots, w_k=w$, then $I_{w,q}$
can be constructed as follows. Start with the interval
$(-\infty,\infty)$, and follow the path downwards.
For each encountered node~$w_i$ splitting on $q$, intersect the current $I_{w,q}$ with either:
\begin{itemize}
    \item $(-\infty,t_{w_i})$ if $w_{i+1}=a_{w_i}$, or
    \item $[t_{w_i},\infty)$ if $w_{i+1}=b_{w_i}$, or
    \item $\emptyset$ otherwise.
\end{itemize}
It follows that the event $X_u=1\land X_v=1$ occurs iff for all
$q\in Q_u\cup Q_v$, $x'_q\in I_{u,q}\cap I_{v,q}$.
Hence, if we denote by $l_{u,v,q},r_{u,v,q}$ the
respective endpoints of $I_{u,q}\cap I_{v,q}$, by the
independence of all perturbations $\delta_j$,
we obtain that $\Pr[X_u=1\land X_v=1]$ equals $\Pi(u,v)$, where
\begin{equation}\label{eq:prod}
\begin{split}
    \Pi(u,v)=\left(\prod\limits_{q\in (Q_u\cup Q_v)\setminus S}[x_q\in I_{u,q}\cap I_{v,q}]\right)\cdot\\
    \prod\limits_{q\in (Q_u\cup Q_v)\cap S}\left(F_{\dist_q}(r_{u,v,q}-x_q)-F_{\dist_q}(l_{u,v,q}-x_q)\right).
    \end{split}
\end{equation}
Implemented naively, computing a single
value $\Pi(u,v)$
takes $O(D_u+D_v)$ time, where $D_u,D_v$
are the depths of $u$ and~$v$ in their respective trees. Indeed, note that
by traversing the root-to-$u$ and root-to-$v$ paths, we can
find all the
relevant intervals $I_{u,q},I_{v,q}$ that the product $\Pi(u,v)$
depends on. Moreover, $\Pi(u,v)$ has at most $D_u+D_v$ factors, each
of which can be evaluated in $O(1)$ time.

Computing the values $\Pi(u,v)$ through all $\Theta(n^2)$ pairs $u,v\in\lvs$ could take
$O(n^2D)$ time, where $D$ is a global bound on the depth of the trees in the ensemble.

However, the computation can be optimized to run in $O(n^2)$ time
by computing the probabilities in an adequate order.
Namely, note that, say, the value $\Pi(a_u,v)$
 can be computed based on $\Pi(u,v)$ in $O(1)$ time.
This is because the interval $I_{a_u,q}$ can differ from the corresponding interval $I_{u,q}$
only for the feature $q=q_u$, and it can be obtained by splitting $I_{u,q}$ with the threshold $t_u$. Hence, only one factor of the product~\eqref{eq:prod} has to be added or replaced in order to obtain $\Pi(a_u,v)$ from $\Pi(u,v)$.
An analogous trick applies to computing
$\Pi(b_u,v)$, $\Pi(u,a_v)$, $\Pi(u,b_v)$
from $\Pi(u,v)$.
Consequently, by iterating through nodes $u$ in pre-order,
and then, once ${u\in \lvs}$ is fixed, through nodes $v$ in pre-order (the order of processing trees does not matter), all the required values $\Pi(u,v)$ can be computed in $O(n^2)$ total time.

This idea is the realized in Algorithm~\ref{alg}, where, while the two nested pre-order traversals over $\tr$ proceed,
the endpoints of intervals $I_{\cdot,\cdot}$ are maintained using the arrays $l,r$,
and the factors of~\eqref{eq:prod} are maintained using the array $\text{factor}$.

\begin{algorithm}[ht!]
\caption{Computing $\Pi(u,v)$ for all leaf pairs $u,v$ given model $\tr$, a feature vector $x$, and important features $S\subseteq [d]$}\label{alg}
\begin{algorithmic}[1]
%\Require{}
\Require{$\forall_{q\in[d]} \quad l[q]=\infty,r[q]=\infty, \text{factor}[q]=1$}
\Procedure{Compute-Prob}{$v,\text{prod}=1,u=\textbf{null}$}
\If{$v$ is non-root}\Comment{update the maintained arrays}
    \State $p:=$ parent of $v$ in $\tr$
    \State $q:=q_p$
    \State $(l',r',f'):=(l[q],r[q],\text{factor}[q])$
    \If{$v=a_p$}\Comment{$v$ is the left child of its parent}
        \State $r[q]:=\min(r[q],t_p)$
    \Else
        \State $l[q]:=\max(l[q],t_p)$
    \EndIf
    \If{$q\in S$}\Comment{See Equation~\eqref{eq:prod}}
        \State $\text{factor}[q]:=F_{\dist_q}(r[q]-x_q)-F_{\dist_q}(l[q]-x_q)$
    \Else
        \State $\text{factor}[q]:=[x_q\geq l[q]\land x_q\leq r[q]]$
    \EndIf
    \State $\text{prod}:=(\text{prod}/f')\cdot \text{factor}[q]$
\EndIf
\If{$v$ is a leaf}
    \If{$u=\textbf{null}$}
        \For{$T\in \tr$}\Comment{inner loop through trees}
            \State \Call{Compute-Prob}{$\text{root}(T),\text{prod},v$}
        \EndFor
    \Else\Comment{both leaves $u$ and $v$ are fixed}
        \State $\Pi(u,v):=\text{prod}$
    \EndIf
\Else\Comment{descend}
    \State \Call{Compute-Prob}{$a_v,\text{prod},u$}
    \State \Call{Compute-Prob}{$b_v,\text{prod},u$}
\EndIf
\If{$v$ is non-root}\Comment{revert changes to $l,r,\text{factor}$}
    \State $(l[q],r[q],\text{factor}[q]):=(l',r',f')$
\EndIf
\EndProcedure
\State
\For{$T\in \tr$}\Comment{outer loop through trees}
    \State \Call{Compute-Prob}{$\text{root}(T)$}
\EndFor
\end{algorithmic}
\end{algorithm}
\begin{theorem}
Let $f$ be a tree ensemble model whose trees have $n$ nodes in total.
Let $x\in \mathbb{R}^d$.
Then for any $S\subseteq [d]$, $\PG^2(x,S)$ can be computed in $O(n^2)$ time.
%In particular, given a feature ranking $\pi$, for any $k\in [d]$, $\PGI^2(x,\pi,k)$ can be computed in $O(n^2)$ time, whereas $\PGI^2(x,\pi)$ can be computed in $O(n^2d)$ time.
\end{theorem}

\section{Experiments: Exact vs Monte Carlo Sampling}

In this section, we compare 
our algorithm computing the squared prediction gap (as defined in Equation~\eqref{eq:pgsquared}) for tree ensembles
%described in Section~\ref{sec:pgi_algorithm},
as described earlier,
the regular Monte Carlo sampling-based method (MC) and Quasi Monte Carlo sampling method (QMC)\footnote{The details of quasi-random generators used in our Quasi Monte Carlo implementation can be found in the Appendix.}. Specifically, the Monte Carlo methods, given $x$ and the set $S\subseteq [d]$ or perturbed features,
evaluates $f(x')$ for~$i$ randomly sampled $x'\sim \text{perturb}(x,S)$
%(see Section~\ref{sec:pgi_algorithm})
and records the average value of $(f(x')-f(x))^2$ over the samples.
We will call $i$ the number of \emph{iterations} in the following.

Recall that the accuracy of Monte Carlo methods is supposed to increase
with the number of iterations $i$. 
The subject of the comparison was hence the accuracy of the results produced by the sampling algorithms as a function of the number of iterations with the goal of deciding which method (exact, MC, or QMC) is preferable and under what conditions.

Specifically, having fixed a model $f$, we looked at the \emph{average difference} between the values $\PG^2(x,S)$ computed using our closed-form algorithm and the two considered sampling methods. Using the output of our algorithm as the ``ground truth'' requires the algorithm to be numerically stable. So an indirect goal of the comparison was to establish whether our algorithm has good numerical properties.

\subsection{Experimental Setup}
\subsubsection{Datasets}\label{subsec:datasets}
Following the previous work (e.g.,~\cite{agarwal2022openxai, dai2022fairness}), in each experiment
we assume that the (real) perturbations used for computing the prediction gap
(either exactly, or approximately)
come from the same normal distribution $\dist=\mathcal{N}(0,\sigma^2)$. This is why we need to assume that all the
features have numerical values.

Using the same perturbation distribution for all the features
requires the features be standardized by subtracting the mean and dividing by the standard deviation. This ensures that if a feature is perturbed by a noise variable drawn from $\dist$, the perturbation will have a similar effect. More specifically, it prevents a perturbation from $\dist$ being relatively significant for one feature and insignificant for another.

In our experiments, we used the following datasets:
\begin{enumerate}
    \item Red Wine Quality \citep{misc_wine_quality_186}. The dataset contains 11 features wine, all numerical and continuous.  The task is to predict the score of a wine, which is an integer between 1 and 10. We considered it as a regression task. The dataset contains $1\ 599$ examples and has 11 features.
    
    \item California Housing  \citep{california_housing_dataset} The dataset contains information from the 1990 Californian census. There are 8 numerical characteristics, and one categorical - proximity to the ocean. For the reasons outlined before, we decided to drop this feature and use a modified dataset.  The task is to predict the median value of the house.  The dataset contains $20\ 640$ examples and has 8 features.

    \item Parkinson Telemonitoring Data
    \citep{misc_parkinsons_telemonitoring_189} The dataset contains $5\ 875$ voice measurements from Parkinson's disease patients, collected at home. It includes 17 numerical features, after dropping 3 categorical columns(ID, age) with a task to predict UPDRS motor and total scores.
\end{enumerate}

\subsubsection{Models} \label{subsec:models}
All tree ensembles are implemented in the XGBoost library \citep{Chen_2016}. For each of the datasets, models of the following two types were trained:
\begin{itemize}
    \item Single tree -- one decision tree with \verb|max_depth| up to~4. 
    \item Bigger -- gradient boosted trees with \verb|max_depth| up to~4 and number of trees limited to 40.
\end{itemize}

In each case, the data set was split 80:20 into training and test sets. In addition, for each model and its constraint, a grid search was performed over selected hyperparameters. The details of the training process can be found in 
the Appendix.
%Appendix~\ref{app:exp1}.

%\subsection{Comparing squared prediction gap algorithms} \label{subsec:comparision}

%\subsubsection{Experimental setup.}
%\adam{przeniesc} to make the comparison fair, the number $i$ was chosen so that the wall time of both algorithms was close.

\subsubsection{Measurements} The comparison
%between the algorithms
was carried out in different settings. For a selected dataset and sampling method (MC/QMC), the variable parameters were:
\begin{itemize}
   \item Model type $m$. 
   %In our case, there were four models in total (Section~\ref{subsec:models}) trained on two datasets (Section~\ref{subsec:datasets}).
   Recall that, in our case, there were two model types for a fixed dataset.
    \item The standard deviation $\sigma$ of a Gaussian used to perturb a feature. %In our experiment,
    We used the values $\{0.1, 0.3, 1.0\}$.
\end{itemize}

To carry out the experiment having fixed the above parameters, we used the following procedure. For each number of iterations ${i\in\{100, 500, 1000, 2000, 4000, 6000, 8000, 10000, 15000}$, $20000, 25000, 30000, 35000\}$, 
we ran our closed-form algorithm and the sampling method in question with iteration count $i$, both estimating the same value $\PG^2(x,S)$ over $N=20\ 000$ random pairs $(x,S)$. For each such pair, we recorded: (1) the average difference between the two estimates and (2) the computation times. 
The $N$ pairs varied in the size of the perturbed feature set
$S$ (which could be, e.g.,  $1,\ldots,11$ for the Red Wine Quality dataset), and we ensured that all subset sizes were equally represented among the pairs.
 Each pair involved a randomly selected point from a test dataset and a random set of features of the desired size.
 %Both  algorithms were then run with appropriate parameters on each prepared sample.

\subsubsection{Implementation} All algorithms were implemented in Python with the key parts including inference on the tree ensemble being implemented as a C++ package and all the arithmetic operations in the algorithms were carried out using the \verb|numpy.float32| type or C++ \verb|float| type. 

The computations were carried out on a FormatServer THOR E221 (Supermicro) server equipped with two AMD EPYC 7702 64-Core processors and 512 GB of RAM with operation system Ubuntu 22.04.1 LTS.

 \subsection{Results}
 %We compared the running times and results of the two algorithms. 
 \subsubsection{Accuracy}
 To estimate the accuracy of sampling-based methods as a function of the iteration count, we computed the Normalised Mean Absolute Error (NMAE), defined as:

\begin{equation*}
\text{NMAE}(y, \hat{y}) = \frac{\sum_{j = 1}^{N} \mid y_j - \hat{y_j}\mid}{\sum_{j = 1}^{N}{\mid y_j \mid}},
\end{equation*}
where $y_j$ is the respective value $\PG(x_j,S_j)$ computed by the exact algorithm, and $\hat{y_j}$ is the respective estimate obtained by the sampling algorithm.
As opposed to the more popular MAPE error, NMAE was used
because it is well defined in cases when true prediction is $0$, which is often the case (for example when the perturbation is small). 

The results obtained in the experiment -- the error as a function of the number of iterations -- are depicted using the ``Mean Monte Carlo'' and ``Mean Quasi Monte Carlo'' curves 
for $\sigma = 0.3$ and Bigger model for the Red Wine Quality dataset in Figure~\ref{fig:results_0.3_wine}.
Analogous figures for other datasets, models and $(m,\sigma)$ can be found in 
the Appendix.
%Appendix~\ref{app:exp1}.
\begin{figure}[!hb]
  \centering
  \includegraphics[width=0.9\linewidth]{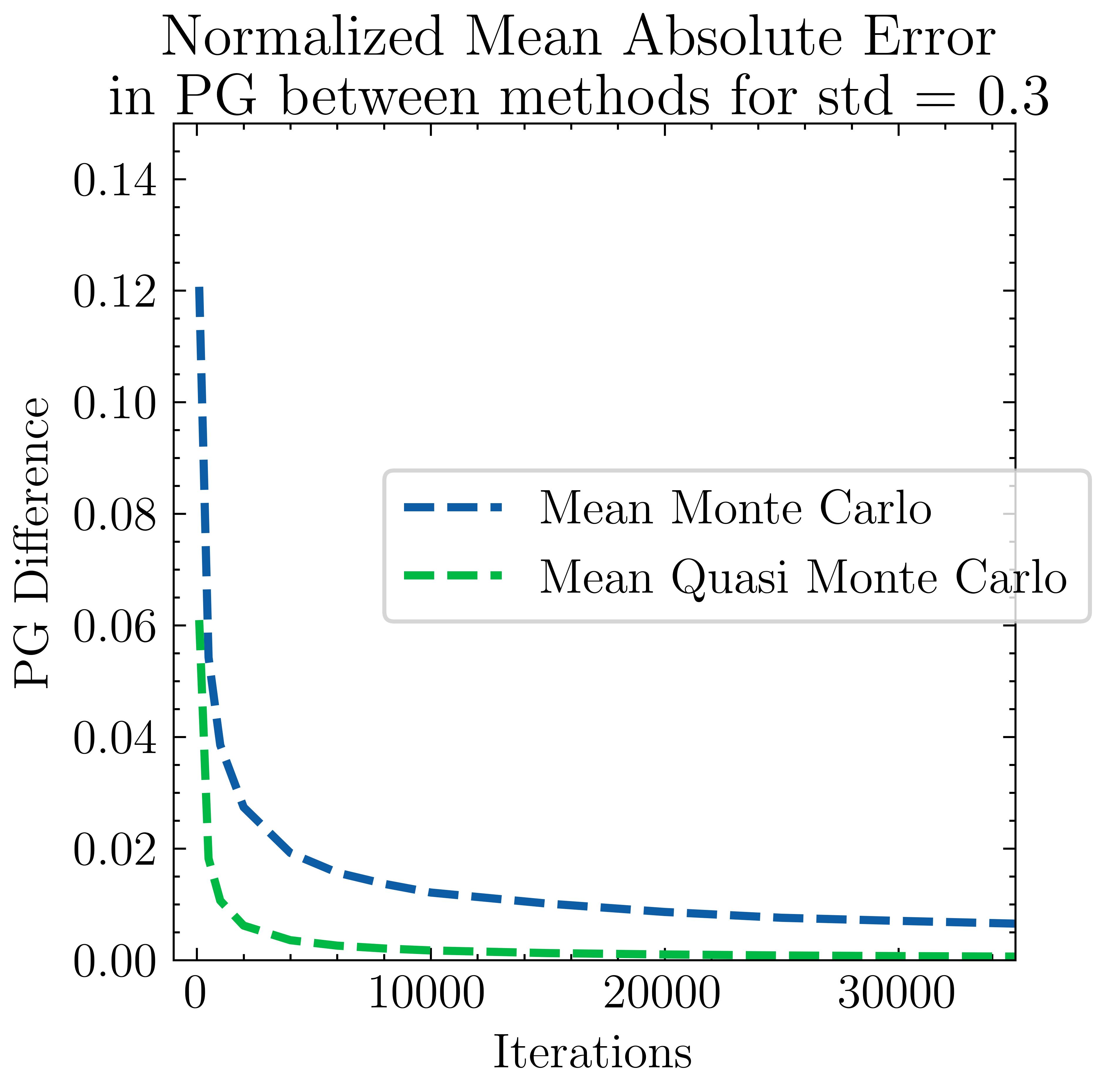}
  \caption{NMAE for $\sigma = 0.3$ and for Bigger model for Red Wine Quality dataset.}\label{fig:results_0.3_wine}
\end{figure}
Overall, the difference between the exact and sampling algorithms decreases as the number of iterations in the sampling algorithm increases. This is indeed anticipated since MC sampling yields better accuracy with iteration increase; in fact, the relative error decreases proportionally to $1/\sqrt{i}$. Additionally, the error for the QMC method is significantly lower, especially for the lower count of iterations.
Moreover, the consistent $\text{NMAE}(y, \hat{y})$ error drop with the increase of the iteration count of the sampling methods indicates that our closed-form algorithm is numerically stable. 
%Additionally, the figures contain separate error curves for some
%selected sizes $|S|$ of the perturbed features set. As can be observed for a given iteration, the NMAE error behaves similarly for all subset sizes. 
%\adam{usunalem to o roznych wielkosciach $|S|$, bo i tak to zniknelo z wykresow, ok?}

\subsubsection{Efficiency} The running time of the sampling methods depends, clearly, linearly on the iterations count.
As a result, the decision about which method to choose should depend on the acceptable accuracy.
To this end, for each of the considered settings (dataset and parameters), we identified the number of iterations (out of $13$ considered thresholds as explained above) in the respective sampling algorithm that came closest to the running time of the exact algorithm for each setting. These corresponding numbers of iterations and the associated NMAEs are shown in Table~\ref{tab:iteration_nmae}.
A larger acceptable NMAE than the corresponding threshold value suggests using the respective sampling-based method, and for smaller NMAEs than that, the exact algorithm is preferable. 
%\adam{jedna tabelka?}
\begin{table}[hb!]
    \centering

    \begin{tabular}{|c|c|c|c|c|}
      %\toprule 
      & \multicolumn{2}{c|}{\bf{Monte Carlo}} & \multicolumn{2}{c|}{\bf{Quasi Monte Carlo}} \\
      \hline
      \bfseries $\sigma$ & Iterations &  NMAE & Iterations  & NMAE \\
      %\midrule  
            \hline

      \multicolumn{5}{c}{Single tree Red Wine Quality model} \\
      \hline
0.1 & 100 & 0.132 & 100 & 0.04\\
0.3 & 100 & 0.136 & 100 & 0.049\\
1.0 & 100 & 0.122 & 100 & 0.054\\
\hline
      \multicolumn{5}{c}{Bigger Red Wine Quality model} \\
            \hline

0.1 & 4000 & 0.019 & 4000 & 0.003\\
0.3 & 8000 & 0.014 & 8000 & 0.002\\
1.0 & 15000 & 0.009 & 15000 & 0.001\\

\hline
      \multicolumn{5}{c}{Single California Housing model} \\
            \hline

0.1 & 100 & 0.158 & 100 & 0.051\\
0.3 & 100 & 0.14 & 100 & 0.043\\
1.0 & 100 & 0.129 & 100 & 0.049\\

\hline
      \multicolumn{5}{c}{Bigger California Housing model} \\
            \hline

0.1 & 4000 & 0.018 & 4000 & 0.002\\
0.3 & 10000 & 0.011 & 10000 & 0.001\\
1.0 & 15000 & 0.01 & 15000 & 0.001\\
\hline
      \multicolumn{5}{c}{Single Parkinson Telemonitoring model} \\
            \hline

0.1 & 500 & 0.071 & 100 & 0.037\\
0.3 & 500 & 0.064 & 500 & 0.011\\
1.0 & 500 & 0.085 & 100 & 0.092\\

\hline
      \multicolumn{5}{c}{Bigger Parkinson Telemonitoring model} \\
            \hline

0.1 & 6000 & 0.015 & 4000 & 0.004\\
0.3 & 15000 & 0.009 & 15000 & 0.002\\
1.0 & 20000 & 0.008 & 20000 & 0.002\\
      \hline  
    \end{tabular}
        \caption{Iterations needed to match the execution time of our exact algorithm, and the corresponding NMAEs.}
        \label{tab:iteration_nmae}
\end{table}

We observe that the vanilla Monte Carlo simulation gives noticeably less accurate results than our exact method with a comparable allocated time budget.
We observe that the error
generally decreases when increasing the size of the model.
This is justified by the fact that our algorithm's running time increases quadratically with $n$, whereas a single iteration of the Monte Carlo and Quasi Monte Carlo method
runs in time at most linear in~$n$. As a result, the bigger the model size is, the more iterations the Monte Carlo and Quasi Monte Carlo methods can perform within the same time budget.
Moreover, for Monte Carlo the error decreases with the standard deviation of the distribution used to obtain perturbations.
%\mg{to już chyba nie bo dla QMC to juz tak nie dziala: This is also expected: for smaller perturbations, the floating-point rounding errors are of higher significance. }
\section{Feature Ranking Using $\PG^2$}
We also study and evaluate a method for ranking the features of a model using the squared prediction gap called \emph{greedy} $\text{PG}^2$.
Let $x\in \mathbb{R}^d$. Suppose the feature perturbations $\delta_i$ all come from $\dist=\mathcal{N}(0,\sigma^2)$. The ranking $\pi_{\PG}^\sigma(x)$ is constructed in the following inductive manner.

%\begin{enumerate}
    %\item For every $i\in [d]$, compute $p_i=\PG^2(x,\{i\})$. Set the most important feature $\pi^\sigma_{\PG}(x)(1)$ to be the feature $i$ for which $p_i$ is the highest.
    %\item
    For $l=0,1,\ldots,d-1$,
    assume we have already computed the $l$ top features $\pi^\sigma_{\PG}(x)(1), \ldots, \pi^\sigma_{\PG}(x)(l)$. Let ${S=\{ \pi^\sigma_{\PG}(x)(1), ..., \pi^\sigma_{\PG}(x)(l)\}}$. For every $i \in [d] \setminus S$, compute $p_i:=\PG^2(x, S\cup \{i\})$. We define the next important feature $\pi^\sigma_{\PG}(x)(l+1)$ to be the feature~$i$ for which $p_i$ is the highest.
%\end{enumerate}

\newcommand{\pishap}{\pi_{\text{SHAP}}}

In the following, we compare the feature rankings obtained using the above method with the rankings obtained using the popular SHAP attribution framework \citep{LundbergECDPNKH20} for tree ensembles, concretely using the \verb|tree_path_dependent| feature perturbation method.

For a given $x\in \mathbb{R}^d$, SHAP produces an attribution vector $(\phi)_{i=1}^d$  whose elements
sum up to $f(x)-\EX[f(x)]$ but can be both positive and negative \citep{SHAP}. 
We define the ranking $\pishap(x)$ 
to be obtained by sorting
the features $[d]$ by the \emph{absolute
value} of $\phi_i$, from highest
to lowest. This is justified
by the fact that the predictions $f(x)$ can be smaller than the expected value;
in such a case, all the attributions $\phi_i$ might be negative.
% 
% \subsection{Degree of agreement between SHAP and $\PG^2$ rankings}

A natural question arises whether this ranking method produces rankings similar or dissimilar to SHAP rankings. Table 2 shows the ratio of datapoints for which the top $k$ most important features are the same for $\pi^\sigma_{\PG}$ and $\pishap$.
As we see, this ratio usually decreases with $k$, increases with~$\sigma$, and is lower for the bigger models than for the single tree models. However, even the rankings for single tree models pick the same most important feature as SHAP less than 76\% of the time. In general, the rankings produced are quite different, so it is worthwhile to study these differences and perform a comparative analysis of the rankings.

\begin{table}[hb!]
    \centering
    
    \begin{tabular}{c|c|c|c}
      %\toprule  
      \bfseries $\sigma$ & $k=1$ &  $k=2$ & $k=3$\\
      %\midrule  
            \hline

      \multicolumn{4}{c}{Single tree Red Wine Quality model} \\
      \hline
0.1 & 0.328 & 0.259 & 0.031 \\
0.3 & 0.412 & 0.591 & 0.428 \\
1.0 & 0.419 & 0.613 & 0.506 \\

\hline
      \multicolumn{4}{c}{Bigger Red Wine Quality model} \\
            \hline

0.1 & 0.131 & 0.041 & 0.022 \\
0.3 & 0.212 & 0.094 & 0.056 \\
1.0 & 0.391 & 0.181 & 0.103 \\

\hline
      \multicolumn{4}{c}{Single California Housing model} \\
            \hline
0.1 & 0.501 & 0.263 & 0.031 \\
0.3 & 0.599 & 0.569 & 0.349 \\
1.0 & 0.712 & 0.471 & 0.504 \\

\hline
      \multicolumn{4}{c}{Bigger California Housing model} \\
            \hline
0.1 & 0.194 & 0.08 & 0.052 \\
0.3 & 0.3 & 0.122 & 0.056 \\
1.0 & 0.569 & 0.243 & 0.08 \\

\hline
      \multicolumn{4}{c}{Single Parkinson Telemonitoring model} \\
            \hline

0.1 & 0.402 & 0.143 & 0.066 \\
0.3 & 0.569 & 0.32 & 0.237 \\
1.0 & 0.759 & 0.414 & 0.22 \\

\hline
      \multicolumn{4}{c}{Bigger Parkinson Telemonitoring model} \\
            \hline
0.1 & 0.137 & 0.028 & 0.007 \\
0.3 & 0.225 & 0.061 & 0.009 \\
1.0 & 0.368 & 0.114 & 0.031 \\
\hline  
    \end{tabular}
    \caption{Ratio of datapoints having the same $k$ most important features in $\pi_{\PG}^\sigma$ as in $\pishap$
    }\label{tab:ranking_comparision}
\end{table}
\subsection{Comparison vs SHAP wrt. $\PGI^2$}
We perform a \emph{global} comparison of $\pi^\sigma_{\PG}$ and $\pishap$ wrt. the $\PGI^2$ metric, that is, we check which ranking
method gives higher $\PGI^2$ scores on average. 
More formally, fix a dataset/model combination $m$ and a perturbation distribution $\dist'=\mathcal{N}(0,(\sigma')^2)$, where potentially $\sigma'\neq\sigma$, for perturbations used
in computing $\PG^2(x,\cdot)$ in~\eqref{eq:pgisquared}.
Let $X\subseteq{\mathbb{R}}^d$ be the corresponding dataset used in $m$. Then, for $\pi\in \{\pishap,\pi_{\PGI}^\sigma\}$, define
the average $\PGI^2$ scores as
\begin{align*}
\overline{{\PGI}^2}(\pi)&=\frac{1}{|X|}\sum_{x\in X}\PGI^2(x,\pi(x)).
\end{align*}
\iffalse
\begin{align*}
\overline{{\PGI}^2}(\pishap)&=\frac{1}{|X|}\sum_{x\in X}\PGI^2(x,\pishap(x)).
\end{align*}
\begin{align*}
\overline{{\PGI}^2}(\pi_{\PGI}^\sigma)&=\frac{1}{|X|}\sum_{x\in X}\PGI^2(x,\pi_{\PGI}^\sigma(x)).
\end{align*}
\fi
Table~\ref{tab:rankings_pgi} presents the 
average scores $\overline{{\PGI}^2}(\pishap)$ and $\overline{{\PGI}^2}(\pi_{\PGI}^\sigma)$ for $\sigma\in\{0.1,0.3,1.0\}$
as a function of $\sigma'$ for Bigger models (an analogous table for Single models can be found in the Appendix). We observe that 
even if the parameter $\sigma$ used when computing the ranking $\pi_{\PG}^\sigma$ differs from the parameter $\sigma'$ used
for computing the $\PGI^2$ metrics, $\overline{{\PGI}^2}(\pi_{\PGI}^\sigma)$
is almost always better than $\overline{{\PGI}^2}(\pishap)$ for the Bigger models, with a slight exception in the ${\sigma'=1.0,\sigma=0.1}$ pair case, when $\sigma,\sigma'$ differ a lot.
%\adam{no nie take zupełnie, gorsze jest dla $\sigma'=1.0$ i $\sigma=0.1$}
%For Single tree models,
%this is the case only for sufficiently large $\sigma'$, but in particular for $\sigma'=\sigma$.
%\adam{moim zdaniem nie ma roznicy Bigger/Single tutaj, sprawdzcie prosze}
\begin{table}[ht!]
\setlength{\tabcolsep}{1mm}
    \centering

    \begin{tabular}{|c|c|c|c|c|}
      %\toprule  
      \bfseries  $\sigma'$ & SHAP & $\sigma$ = $0.1$ & $\sigma$ = $0.3$& $\sigma$ = $1.0$\\
      %\midrule  
\hline
      \multicolumn{5}{c}{Bigger PGI Red Wine Quality model} \\
            \hline
0.1 & 0.013 & 0.024 & 0.022 & 0.018 \\
0.3 & 0.051 & 0.068 & 0.078 & 0.067 \\
1.0 & 0.221 & 0.192 & 0.266 & 0.302 \\

\hline
      \multicolumn{5}{c}{Bigger PGI California Housing model} \\
            \hline
0.1 & 3.346e+08 & 4.205e+08 & 3.841e+08 & 3.065e+08 \\
0.3 & 1.209e+09 & 1.297e+09 & 1.374e+09 & 1.259e+09 \\
1.0 & 7.807e+09 & 6.963e+09 & 8.266e+09 & 8.728e+09 \\

\hline
      \multicolumn{5}{c}{Bigger PGI Parkinson Telemonitoring model} \\
            \hline
0.1 & 2.743 & 4.425 & 4.015 & 3.199 \\
0.3 & 6.341 & 9.458 & 11.878 & 10.143 \\
1.0 & 16.298 & 19.56 & 30.188 & 34.013 \\
      \hline
    \end{tabular}
        \caption{Greedy $\PG^2$ vs SHAP rankings wrt. $\PGI^2$.}\label{tab:rankings_pgi}
\end{table}

% \begin{table}
%     \centering
%     \caption{PGI housing model}\label{tab:pgi_housing}
%     \begin{tabular}{|c|c|c|c|c|}
%       \toprule  
%       \bfseries Ranking $\sigma$ & SHAP & PGI $\sigma$ = $0.1$ & PGI $\sigma$ = $0.3$& PGI $\sigma$ = $1.0$\\
%       \midrule  
% 0.1 & 3.346e+08 & 4.205e+08 & 3.841e+08 & 3.065e+08 \\
% 0.3 & 1.209e+09 & 1.297e+09 & 1.374e+09 & 1.259e+09 \\
% 1.0 & 7.807e+09 & 6.963e+09 & 8.266e+09 & 8.728e+09 \\
%       \bottomrule  
%     \end{tabular}
% \end{table}
% \begin{table}
%     \centering
%     \caption{PGI Parkinson Telemonitoring  model}\label{tab:pgi_telemetry}
%     \begin{tabular}{|c|c|c|c|c|}
%       \toprule  
%       \bfseries Ranking $\sigma$ & SHAP & PGI $\sigma$ = $0.1$ & PGI $\sigma$ = $0.3$& PGI $\sigma$ = $1.0$\\
%       \midrule  
% 0.1 & 2.743 & 4.425 & 4.015 & 3.199 \\
% 0.3 & 6.341 & 9.458 & 11.878 & 10.143 \\
% 1.0 & 16.298 & 19.56 & 30.188 & 34.013 \\
%       \bottomrule  
%     \end{tabular}
% \end{table}

We conclude that the greedy optimization of the $\PGI^2(x,\pi)$ metric performs reasonably well, even though its correctness could only be guaranteed if it optimized the $\PGI(x,\{\pi_1\})$ metric (where only the single most important feature is perturbed) instead of~\eqref{eq:pgisquared}.
%\vspace{-1mm}
\subsection{Comparison vs SHAP with Feature Removing}
We compare the rankings $\pishap$ and $\pi_{\PG}^\sigma$ using feature removing. Intuitively, removing information about the value of a highly important feature should result in greater model error than removing information about the value of a less important feature.
Thus, we can assess a ranking by measuring the error of the model when the features the ranking deems most important are removed. More precisely, for a ranking $\pi$ and for each $k=1,2,...$ we measure the RMSE of $\xi(x, [d] \setminus \pi(x)[1..k])$ over all $x$ from the test dataset, where $\xi(\cdot, S)$ is a function giving model predictions based only on features from $S$. %\rapicki{work in progress from here} 
We have two such functions $\xi$ (and thus two different methods of comparing rankings) defined below.
%There are various ways of defining such a function $\xi$, e.g. retraining the model with certain features removed \citep{HookerEKK19}, replacing the removed features with a certain average value [citation??], or calculating the expected value of the model's prediction when the removed features are randomised.

%The approach we apply utilises feature randomising. We define $ \xi(x, S) := \mathbb{E} [m(X)] $ where $m$ is the model and $X$ is a random variable with $X_i \equiv x_i$ for $i\in S$ and $X_j$ has the distribution of the $j$-th feature over the test set for $j\not\in S$. In practice, the expected value is approximated by sampling $m(X)$ 100 times and taking the average.

The first method is feature randomising. We define $ \xi_{random}(x, S) := \mathbb{E} [m(X)] $ where $m$ is the model and $X$ is a random variable with $X_i \equiv x_i$ for $i\in S$ and $X_j$ has the distribution of the $j$-th feature over the test set for $j\not\in S$. In practice, the expected value is approximated by sampling $m(X)$ 100 times (where each feature is sampled independently) and taking the average.%\adam{czyli kazdy feature spoza $S$ jest samplowany niezaleznie?} \rapicki{Niestety tak. W literaturze spotkalem podejscie ze nieznane features powinny miec rozklad warunkowany wartosciami znanych features, ale to juz bylaby gleboka rozkmina. }

The other method follows the remove-and-retrain approach described in \cite{HookerEKK19}. For $k=1,2,3,...$ and for each subset of features $S \subset [d]$ such that $|S| = d-k$ we train a XGBoost model where we only use the features in $S$. That is, the $k$ features that are not in $S$ are not available for the training of the model. This way, we obtain a family of models $(m_S)_{S \subset [d]}$ where each model $m_S$ gives predictions based only on values of features in $S$. We define $\xi_{retrain}(x, S) := m_S(x)$.

The comparison was performed for $k=1,2,3$. The results are shown in Tables in the Appendix. 
The results for the retraining method suggest that $\pishap$ picks the most important features more accurately than $\pi_{\PG}^\sigma$.
The results for the randomising method show no significant difference between $\pishap$ and $\pi_{\PG}^\sigma$ for Red Wine Quality and Parkinson Telemonitoring models. However, for the Housing model we see that $\pi_{\PG}^\sigma$ with $\sigma \in \{0.1, 0.3\}$ give better results than $\pishap$.
%In other words, the top $k$ features found by greedy $\PG^2$ with $\sigma \in \{0.1, 0.3\}$ seem to be on average more important for the model than the top $k$ features found by SHAP, for $k=1,2,3$.
In general, this comparison doesn't give a simple conclusion that one ranking method is clearly better than the other. Each of them can show advantage over the other depending on the dataset and assessment method.
% For Red Wine Quality and Parkinson Telemonitoring models, all rankings have very close results, the differences are within the confidence intervals. For Housing model, the differences between the rankings are statistically significant. The rankings $\pi_{\PG}^\sigma$ with $\sigma \in \{0.1, 0.3\}$ give better results than $\pishap$, while $\pi_{\PG}^{1.0}$ performs worse than $\pishap$. In other words, the top $k$ features found by greedy $\PG^2$ with $\sigma \in \{0.1, 0.3\}$ seem to be on average more important for the model than the top $k$ features found by SHAP, for $k=1,2,3$.

\section{Conclusion}
In this work, we proposed a cleanly defined faithfulness metric $\PGI^2$ that can be computed exactly in polynomial time on tree ensemble models. We designed a quadratic algorithm for that. Our experimental evaluation showed that the algorithm is numerically stable and superior to Monte Carlo-based methods if very accurate results are desired.

We also proposed a natural feature ranking method inspired by $\PGI^2$ optimization. The method generally identifies different most important features than the state-of-the-art SHAP explainer while offering similar performance wrt. different faithfulness metrics. The $\PG^2$-based ranking may thus offer a viable alternative for assessing feature importance in tree-based models.

\section{Acknowledgments}
Partially supported by the ERC PoC grant EXALT no 101082299 and the National Science Centre (NCN) grant no. 2020/37/B/ST6/04179.

\bibliography{references}
\appendix
\onecolumn

\newcommand{\twocharts}[2] {\includegraphics[width=0.48\textwidth]{#1}%
\hfill%
\includegraphics[width=0.48\textwidth]{#2}%
}

\section{Models training} \label{app:training}

The grid search was performed by using class $GridSearchCV$ from library XGBoost \cite{Chen_2016} to minimize squared error, with numbers of folds in cross-validation set to $5$. Additionally to parameters described in \ref{subsec:models}, the grid search included also parameters with the following possible values:
\begin{itemize}
    \item $max\_depth$ with possible values in $\{1, 2, 3, 4\}$
    \item $eta$ with possible values in $\{0.01, 0.1, 0.2, 0.3, 0.4, 0.5, 0.6, 0.7, 0.8, 0.9\}$
    \item $subsample$ with possible values in ${0.01, 0.1, 0.2, 0.3, 0.4, 0.5, 0.6, 0.7, 0.8, 0.9}$
\end{itemize}
The chosen parameters for models were:
\begin{itemize}
    \item  Single tree model for Wine Quality dataset -- $eta=0.9$, $subsample=0.9$and $max\_depth=4$
    \item  Bigger model for Wine Quality dataset -- $eta=0.2$, $subsample=0.9$, $max\_depth=4$ and $40$ trees
    \item  Single tree model for Californian Housing dataset -- $eta=0.9$, $subsample=0.9$ and $max\_depth=3$
    \item  Bigger model for Californian Housing dataset -- $eta=0.2$, $subsample=0.9$, $max\_depth=4$ $40$ trees
    \item  Single tree model for Parkinson Telemonitoring dataset -- $eta=0.9$ and $subsample=0.9$, $max\_depth=3$
    \item  Bigger model for Parkinson Telemonitoring dataset -- $eta=0.3$, $subsample=0.8$ and $max\_depth=4$
\end{itemize}
\section{Quasi Monte Carlo}

In our implementation of the Quasi-Monte Carlo method, we utilized the quasi-random generator provided by the SciPy library~\citep{2020SciPy-NMeth}. Specifically, we employed the Halton sequence~\citep{halton1960efficiency} for generating our quasi-random points. This choice was motivated by the Halton sequence's flexibility, as it does not necessitate a sample size that is a power of two $(2^n)$, being a better fit for experimental designs. To transform the uniform distribution produced by the Halton sequence into the required normal distribution, we applied the inverse transform method. It is worth noting that while this transformation is effective, it does result in some loss of the low-discrepancy properties inherent to the original Halton sequence.

\section{Additional experiments results} \label{app:exp1}
\subsection{Results from precision experiments}

\begin{figure}[!htb]
    \twocharts{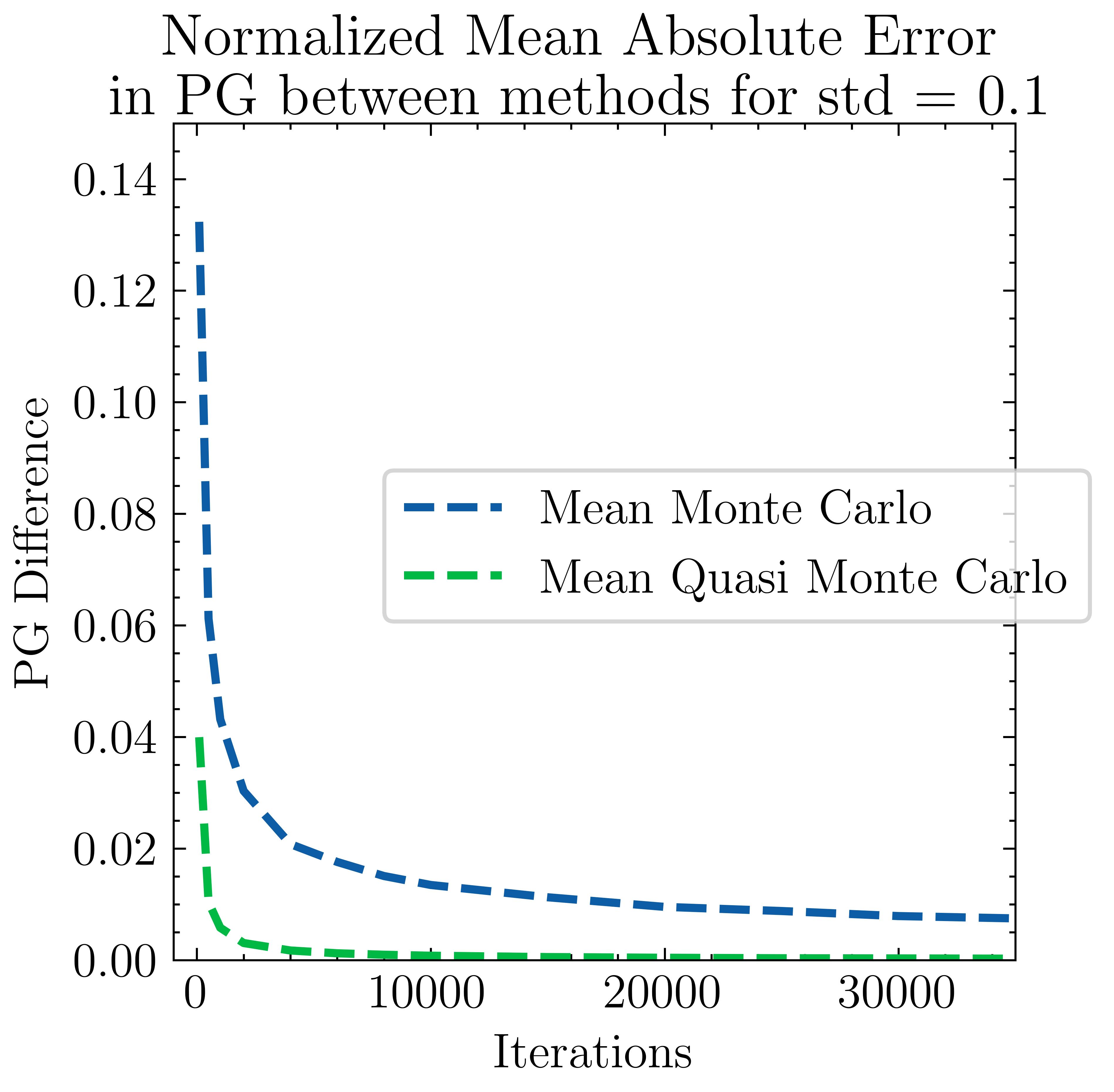}{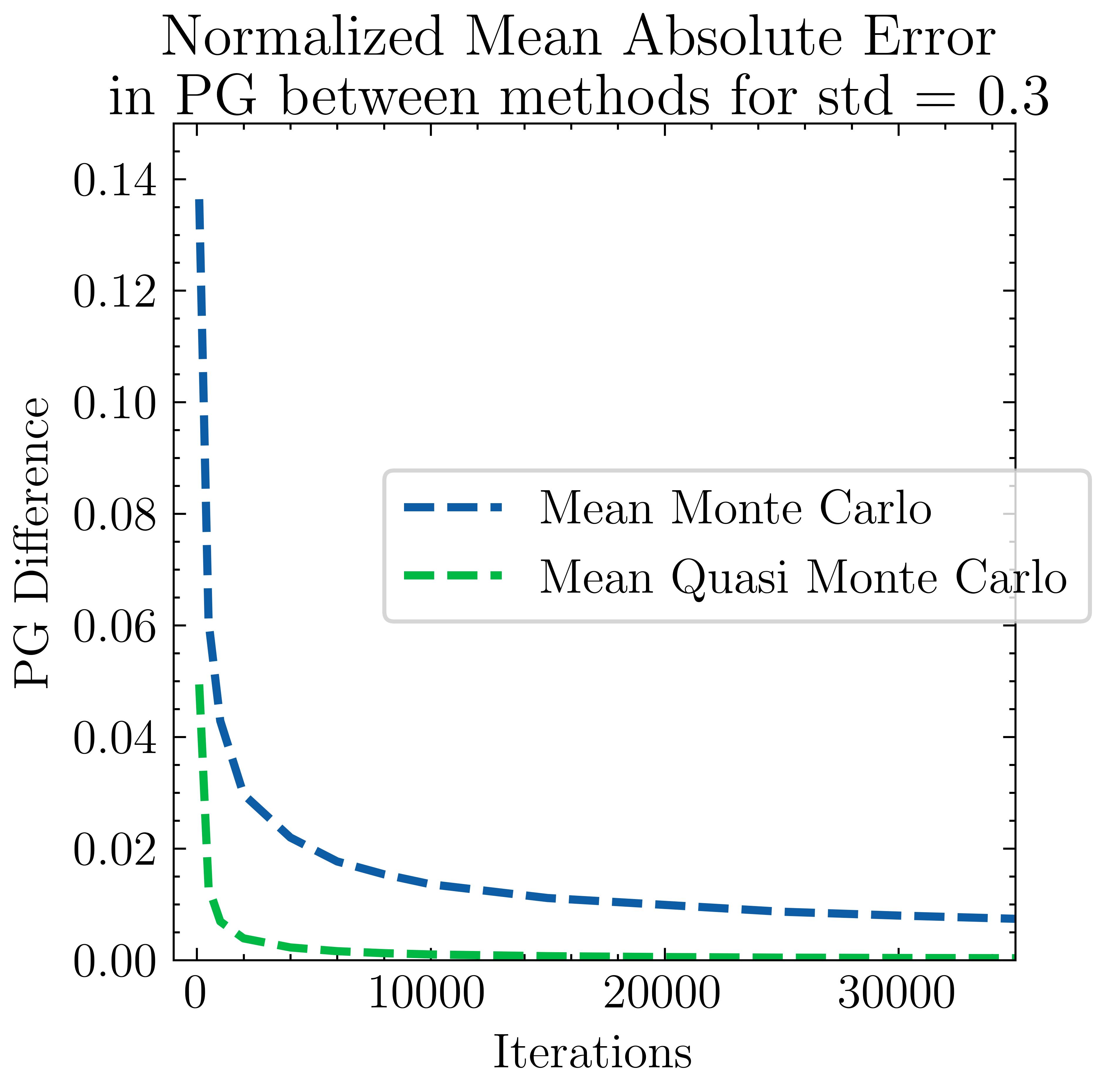}
    \includegraphics[width=0.48\textwidth]{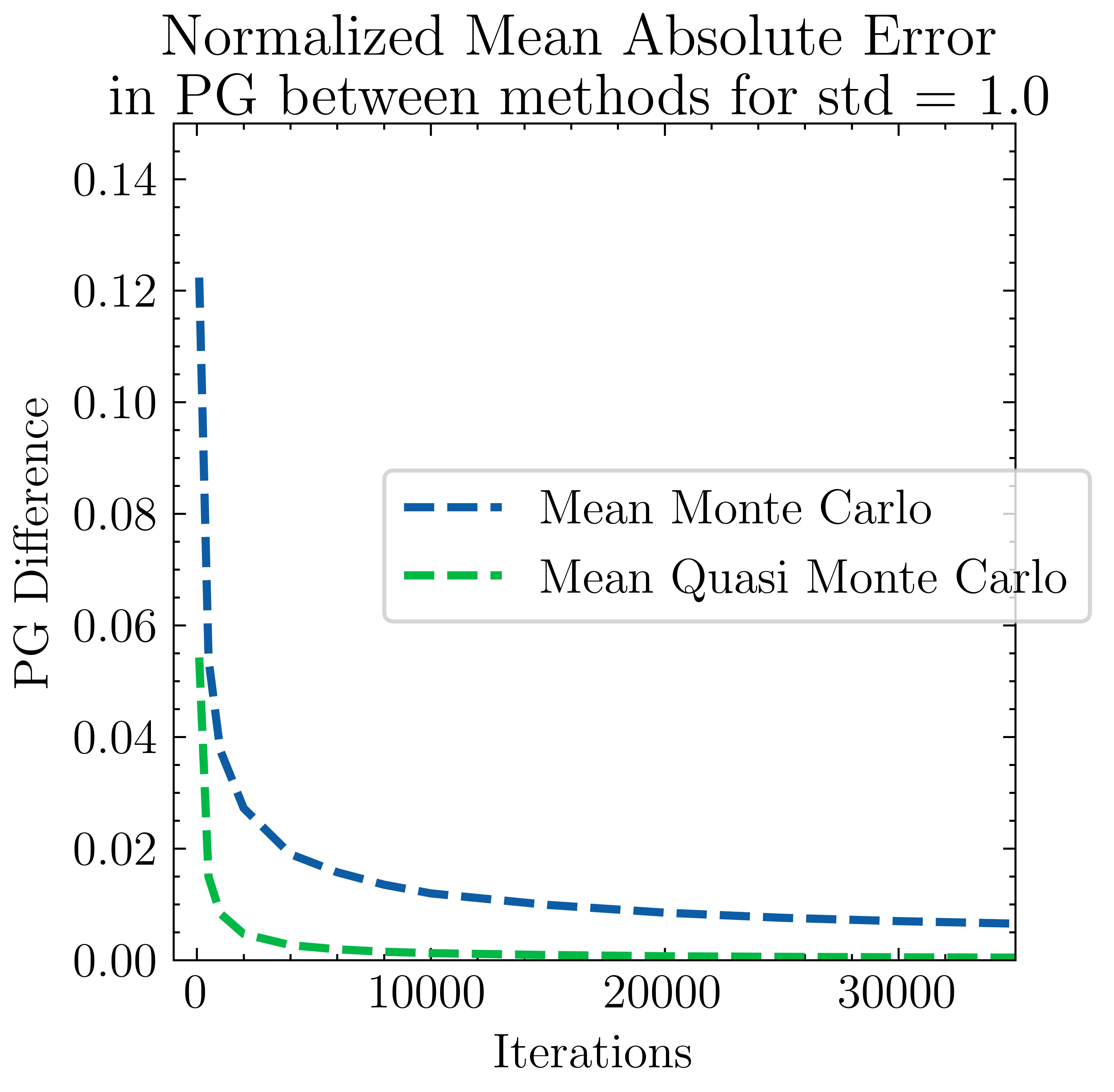}
    \caption{Single tree model for Wine Quality dataset}
    \label{fig:}
\end{figure}

\begin{figure}[!htb]
    \twocharts{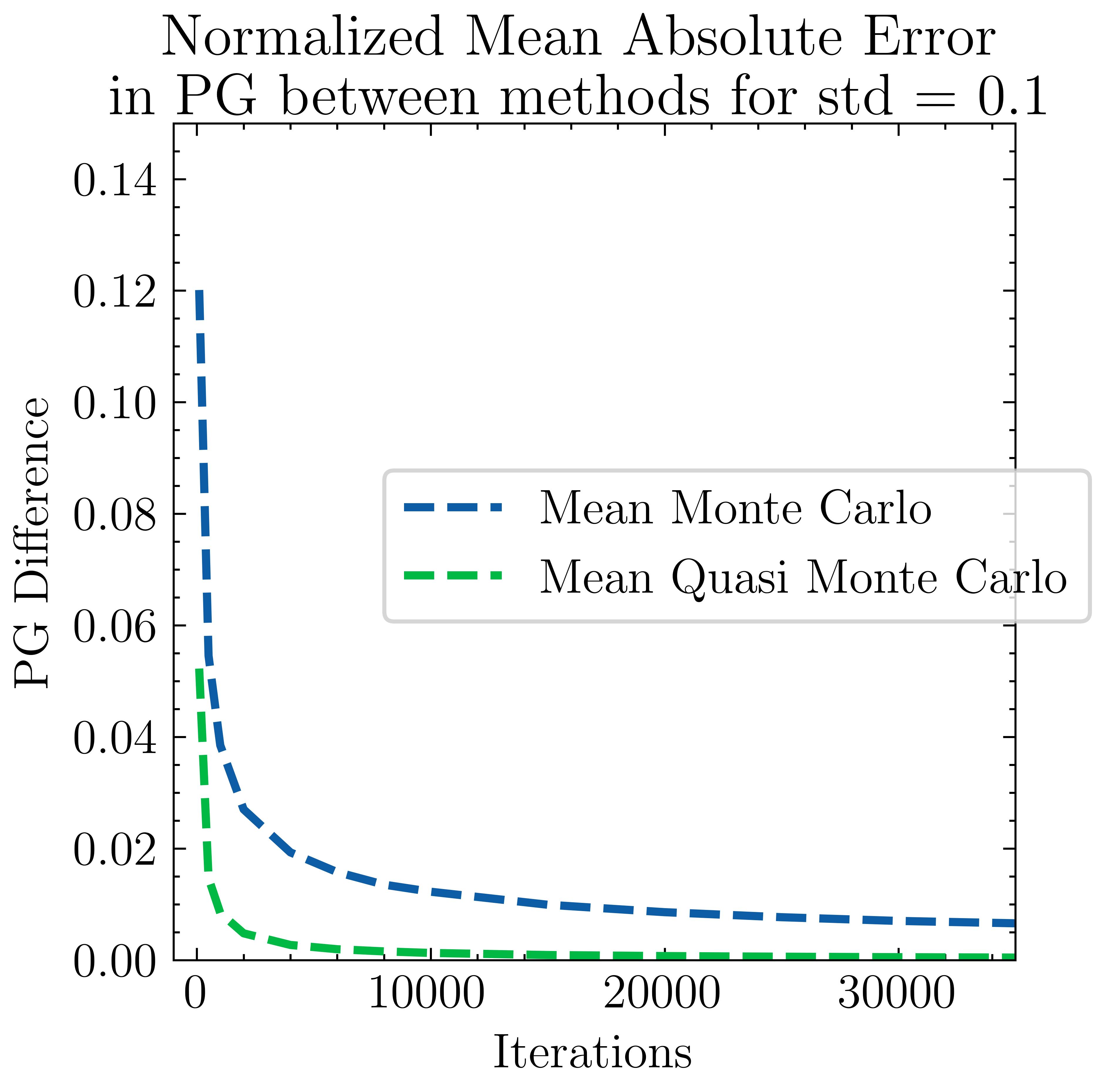}{figures/precision_wine_model_0.3.jpg}
    \includegraphics[width=0.48\textwidth]{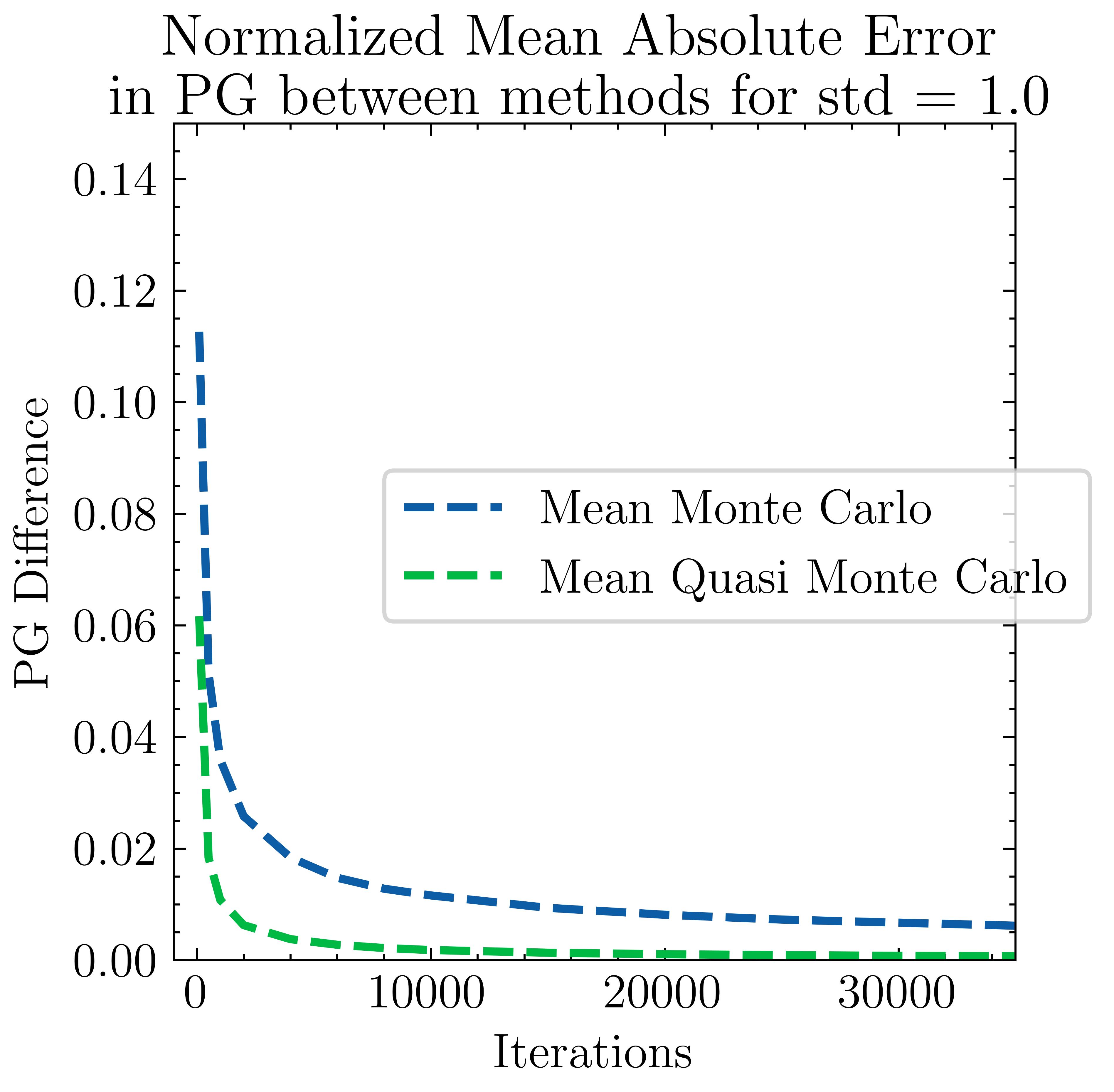}
        \caption{Bigger model for Wine Quality dataset}
    \label{fig:}
\end{figure}

\begin{figure}[!htb]
    \twocharts{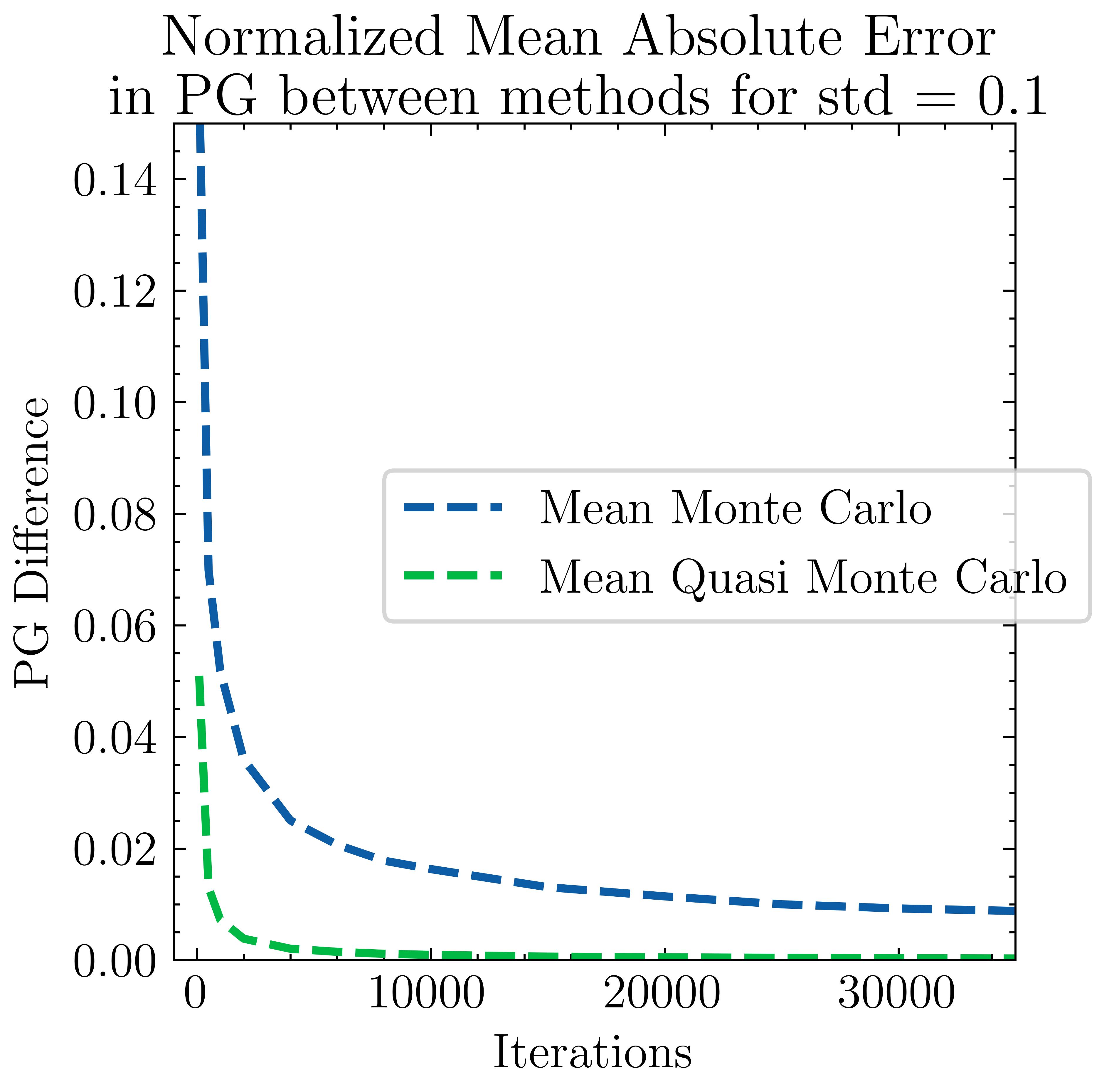}{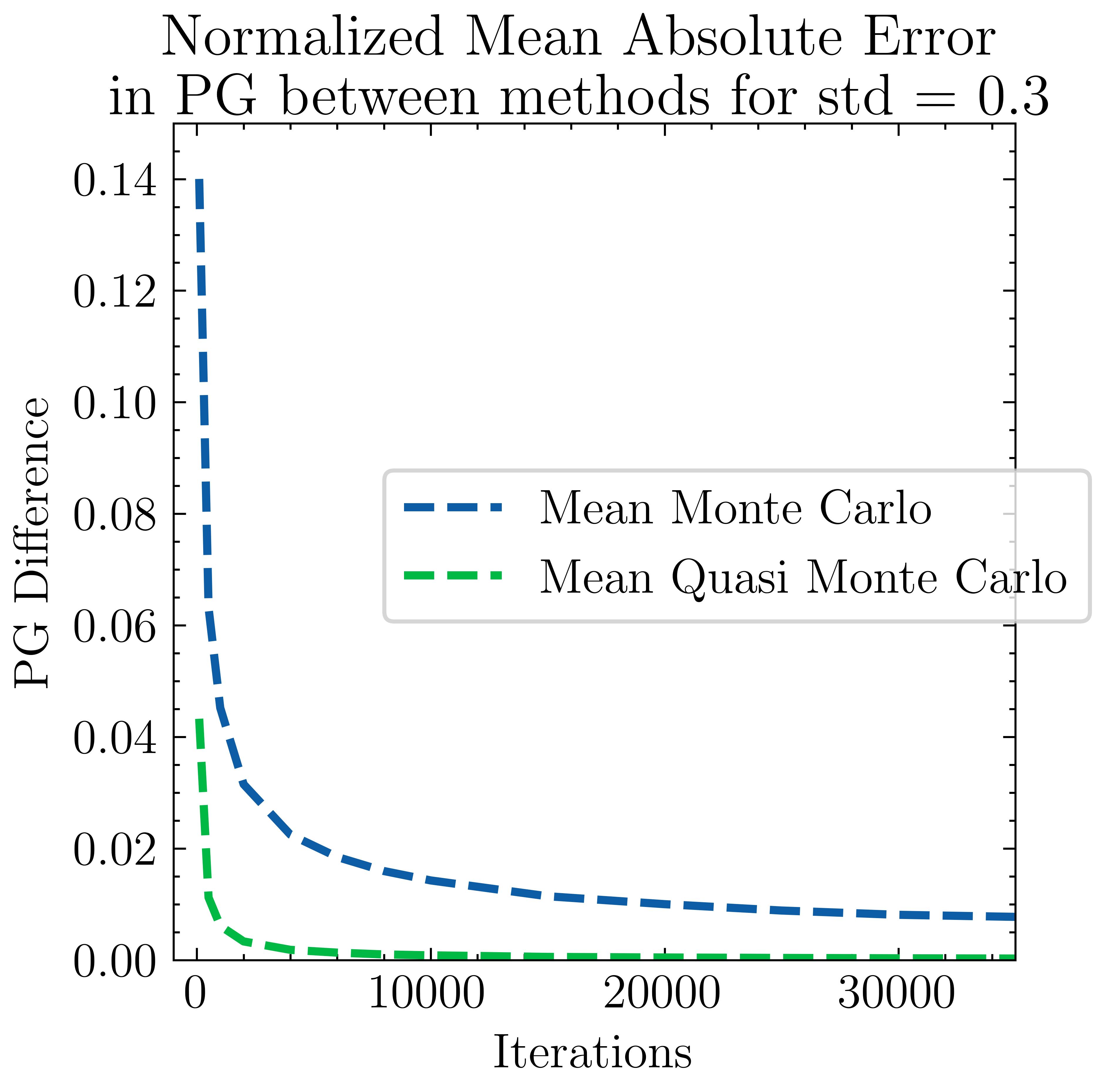}
    \includegraphics[width=0.48\textwidth]{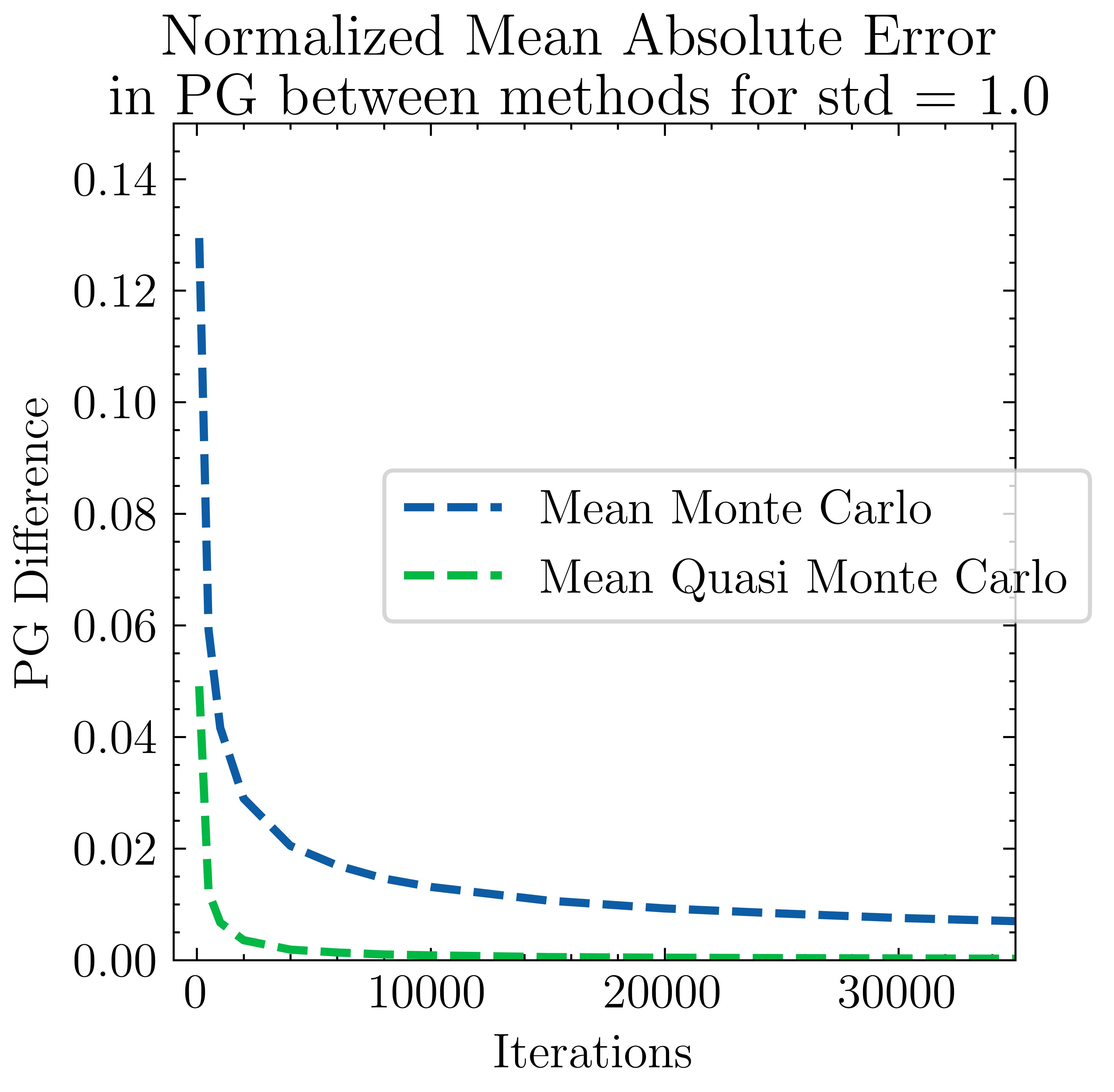}
    \caption{Single tree model for Californian Housing dataset}
    \label{fig:}
\end{figure}

\begin{figure}[!htb]
    \twocharts{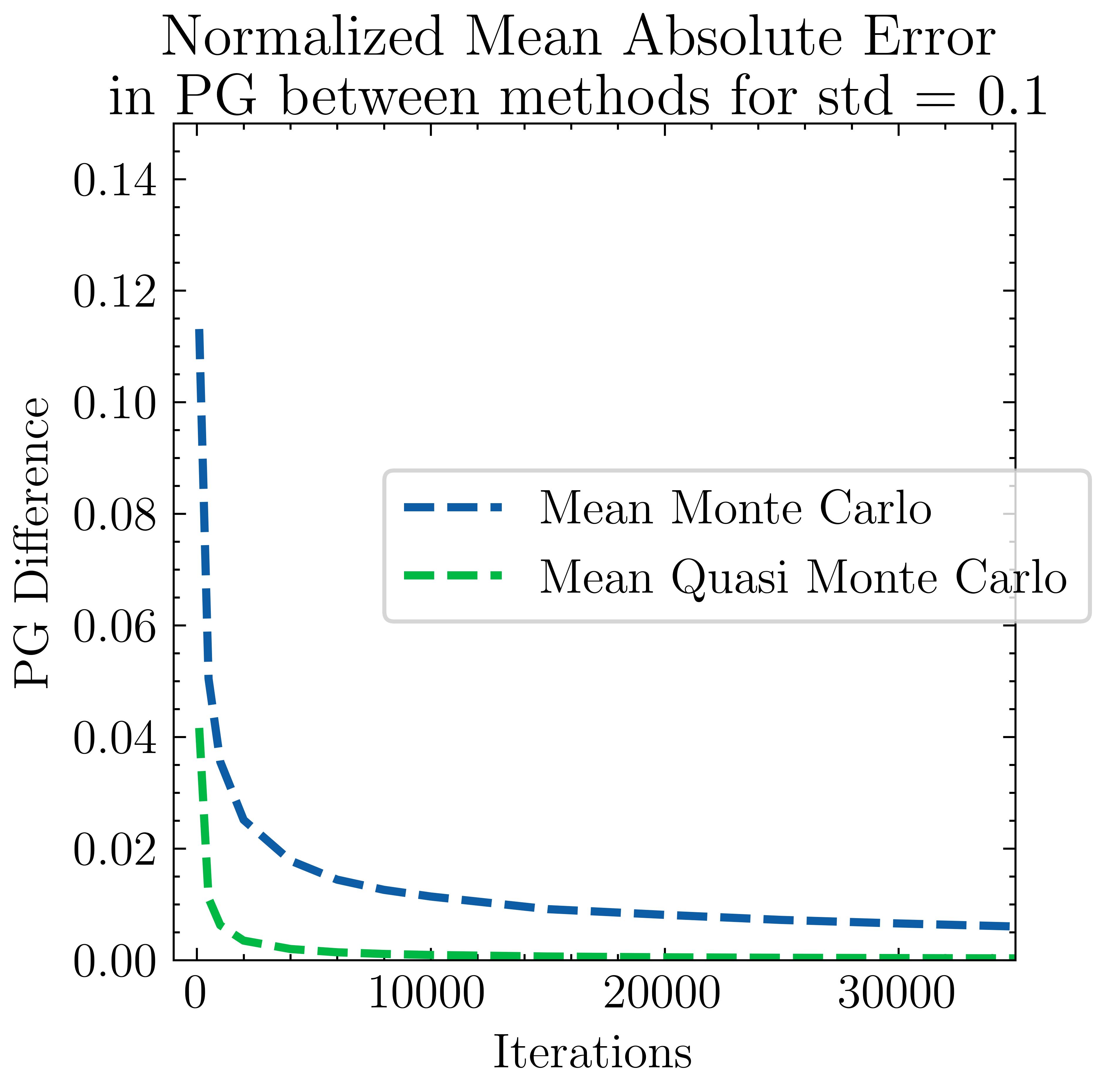}{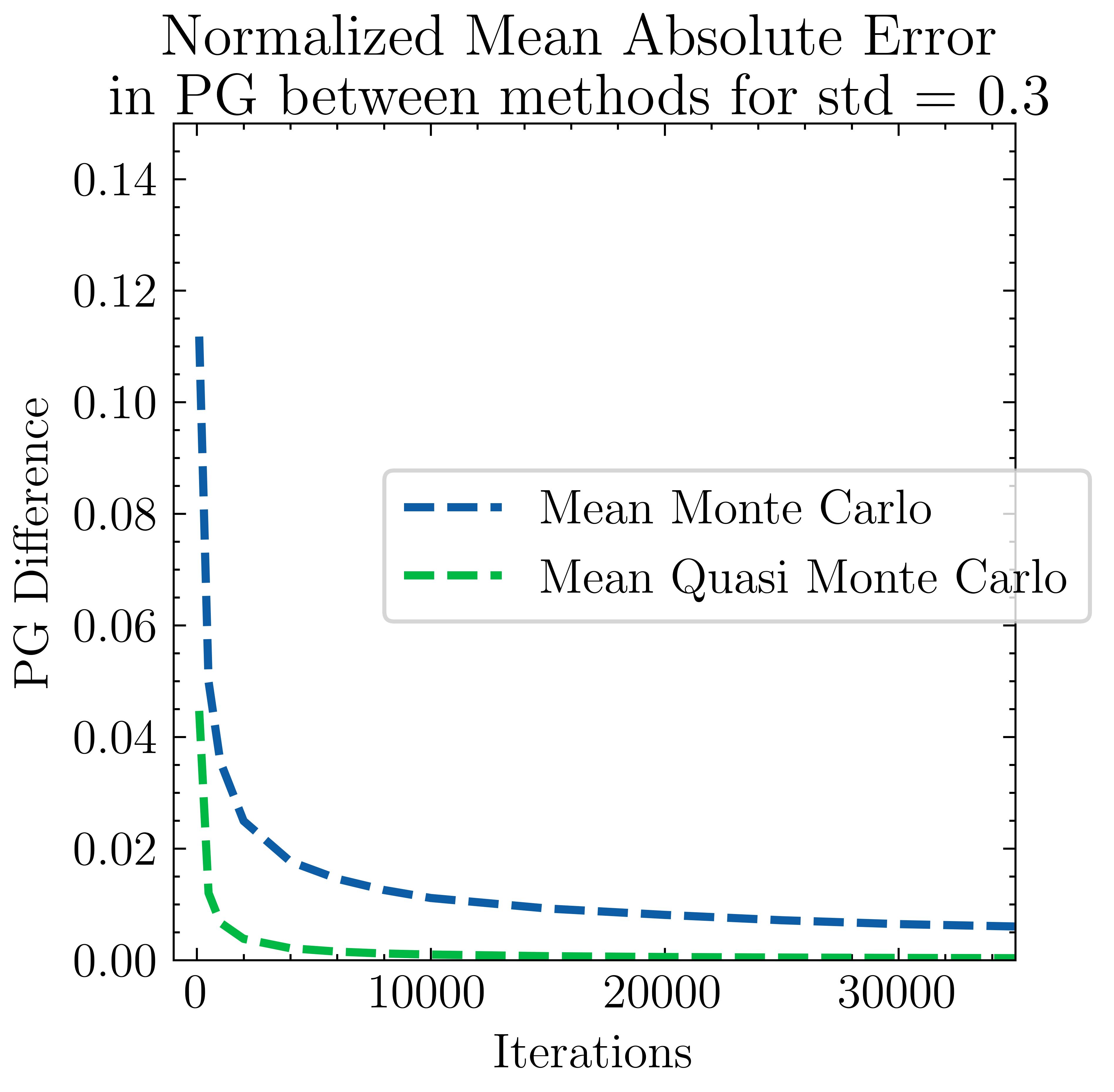}
    \includegraphics[width=0.48\textwidth]{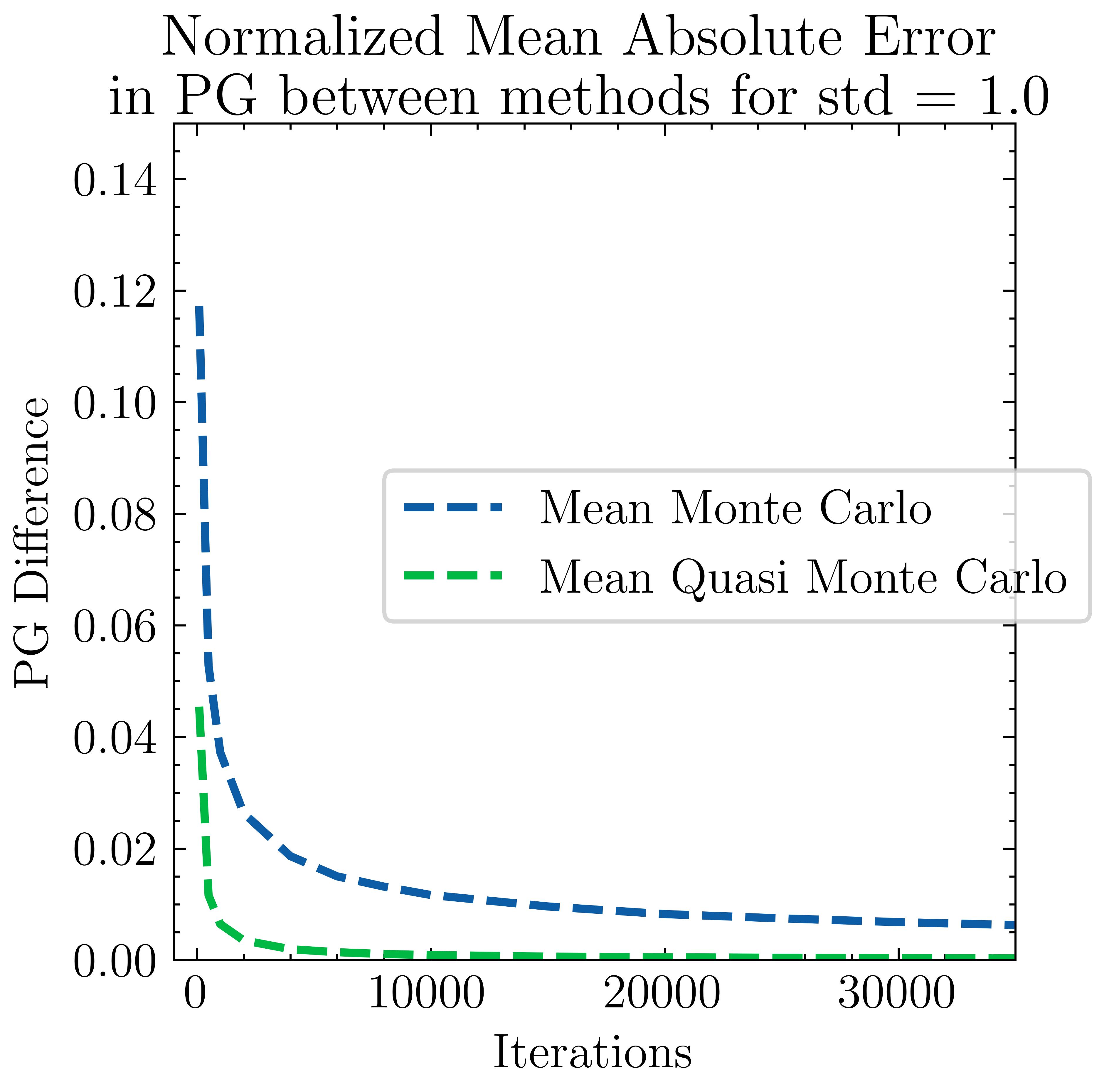}
    \caption{Bigger model for Californian Housing dataset}
    \label{fig:}
\end{figure}

\begin{figure}[!htb]
    \twocharts{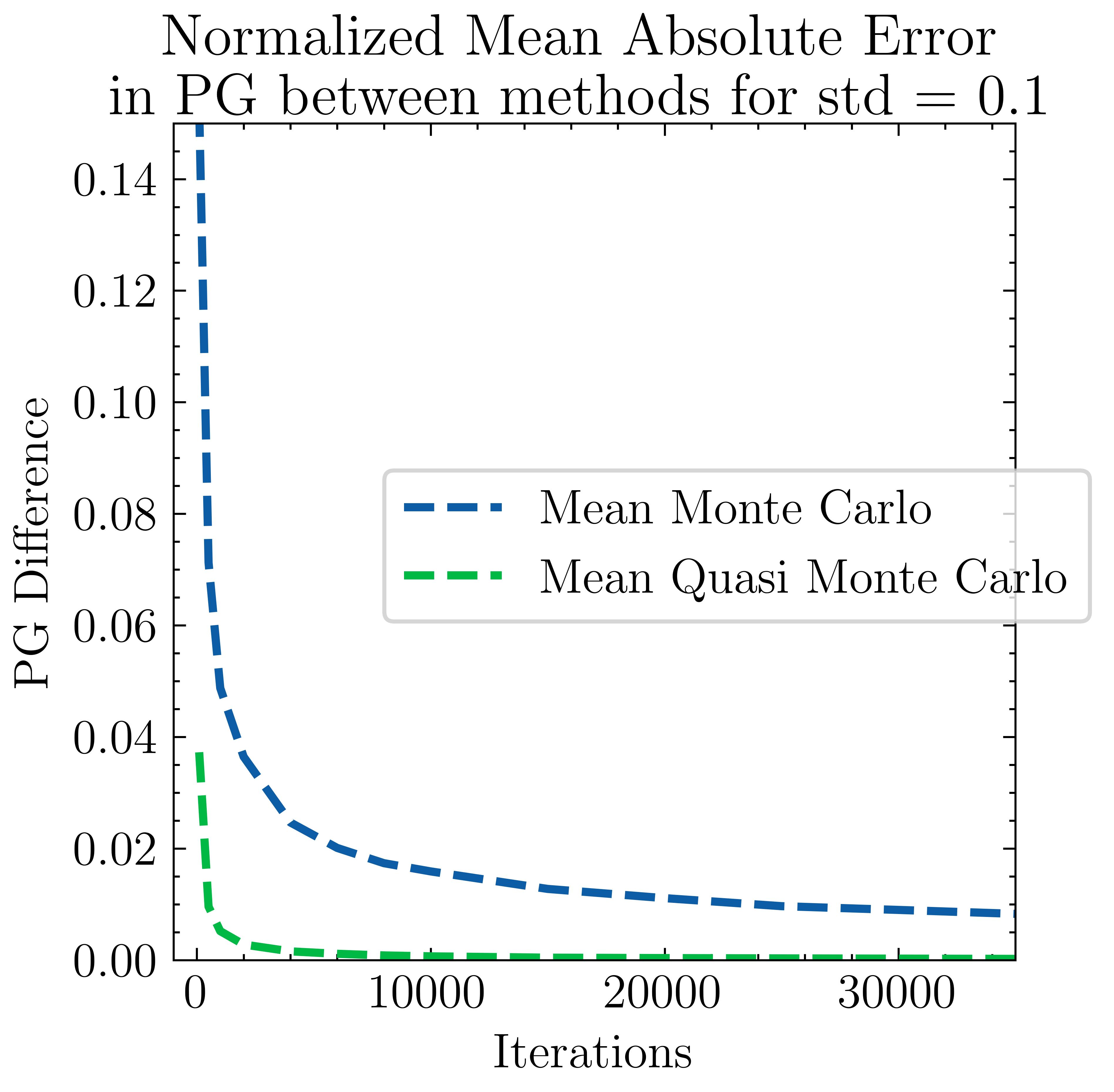}{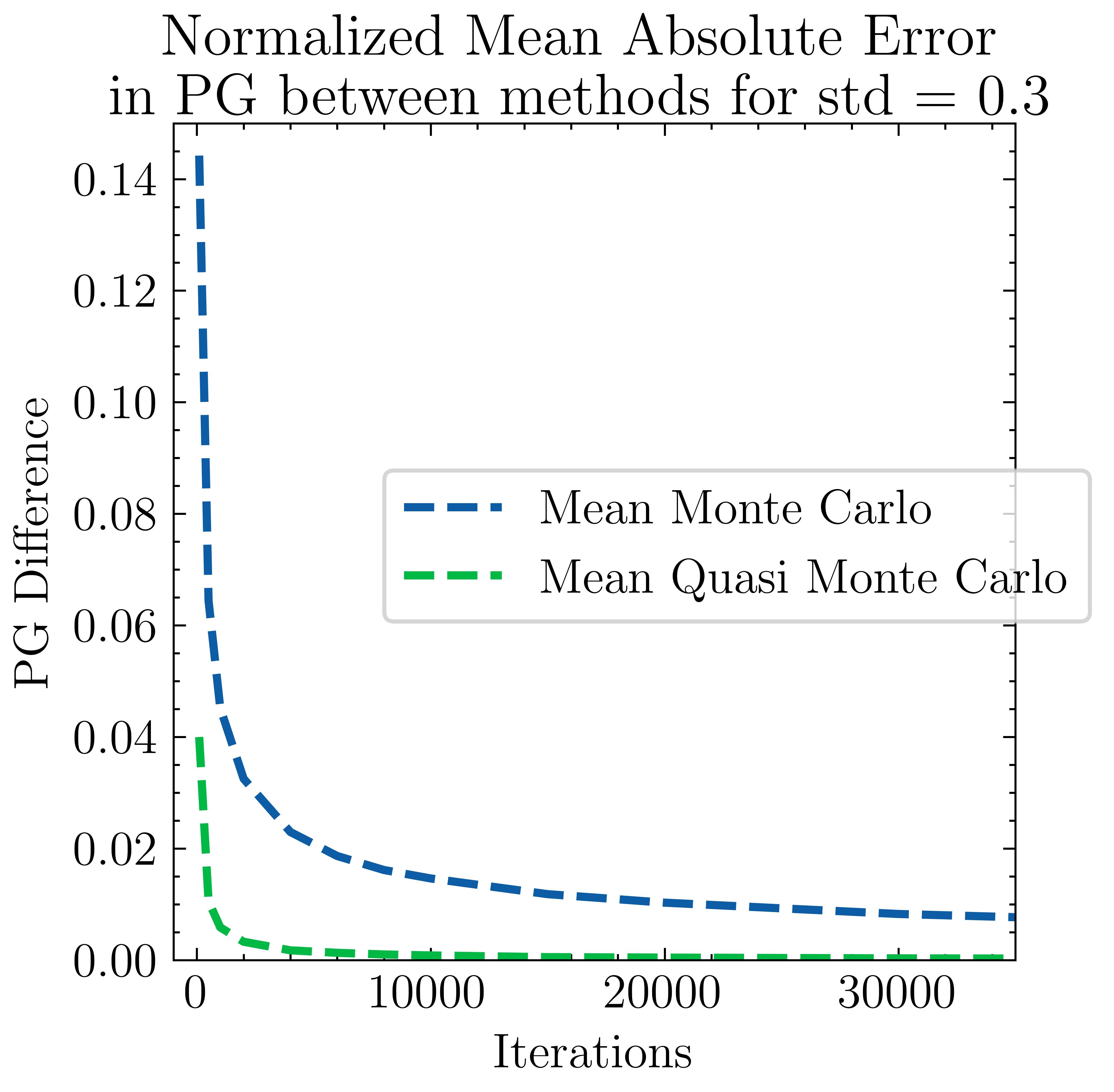}
    \includegraphics[width=0.48\textwidth]{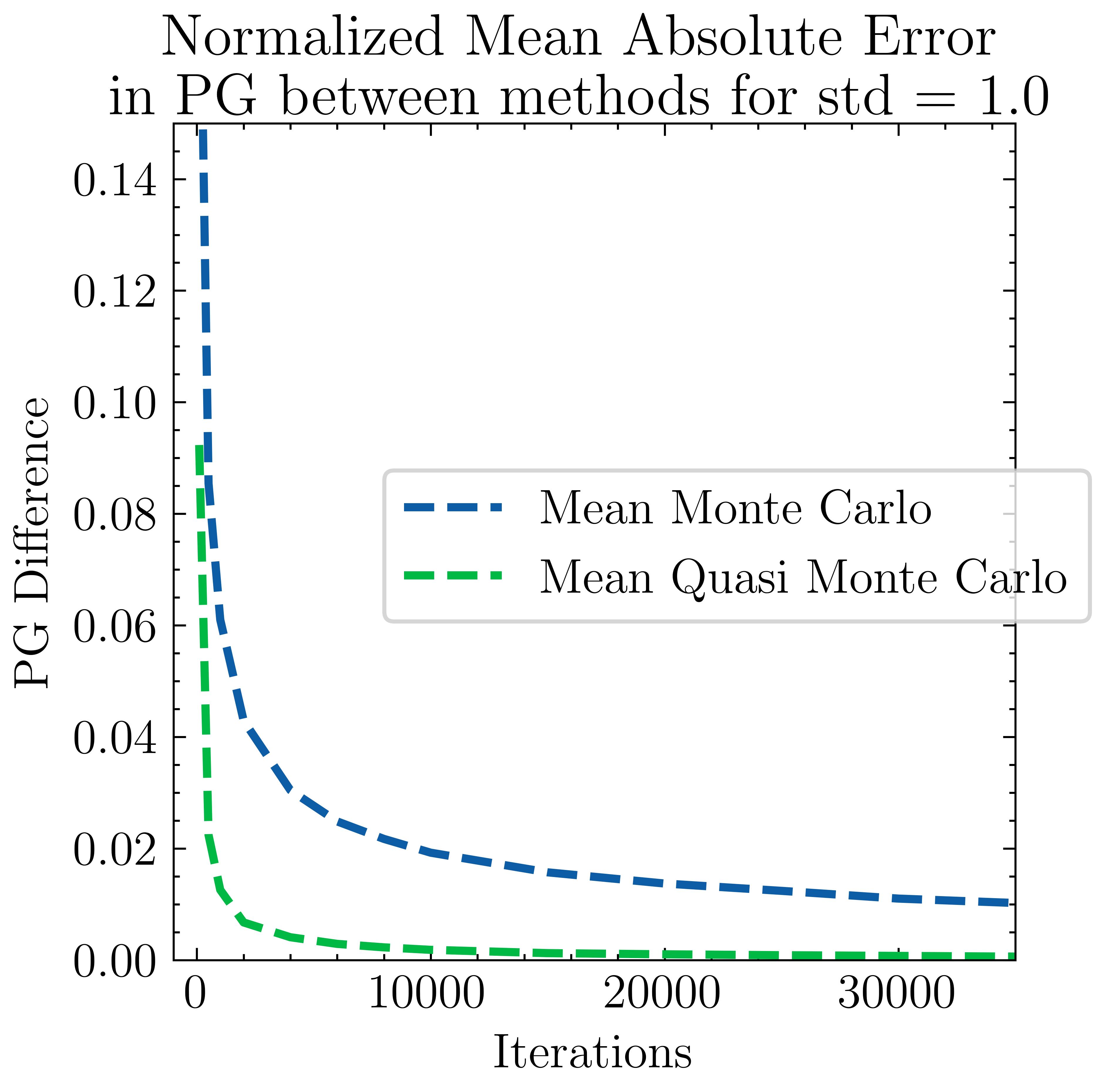}
    \caption{Single tree model for Parkinson Telemonitoring  dataset}
    \label{fig:}
\end{figure}

\begin{figure}[!htb]
    \twocharts{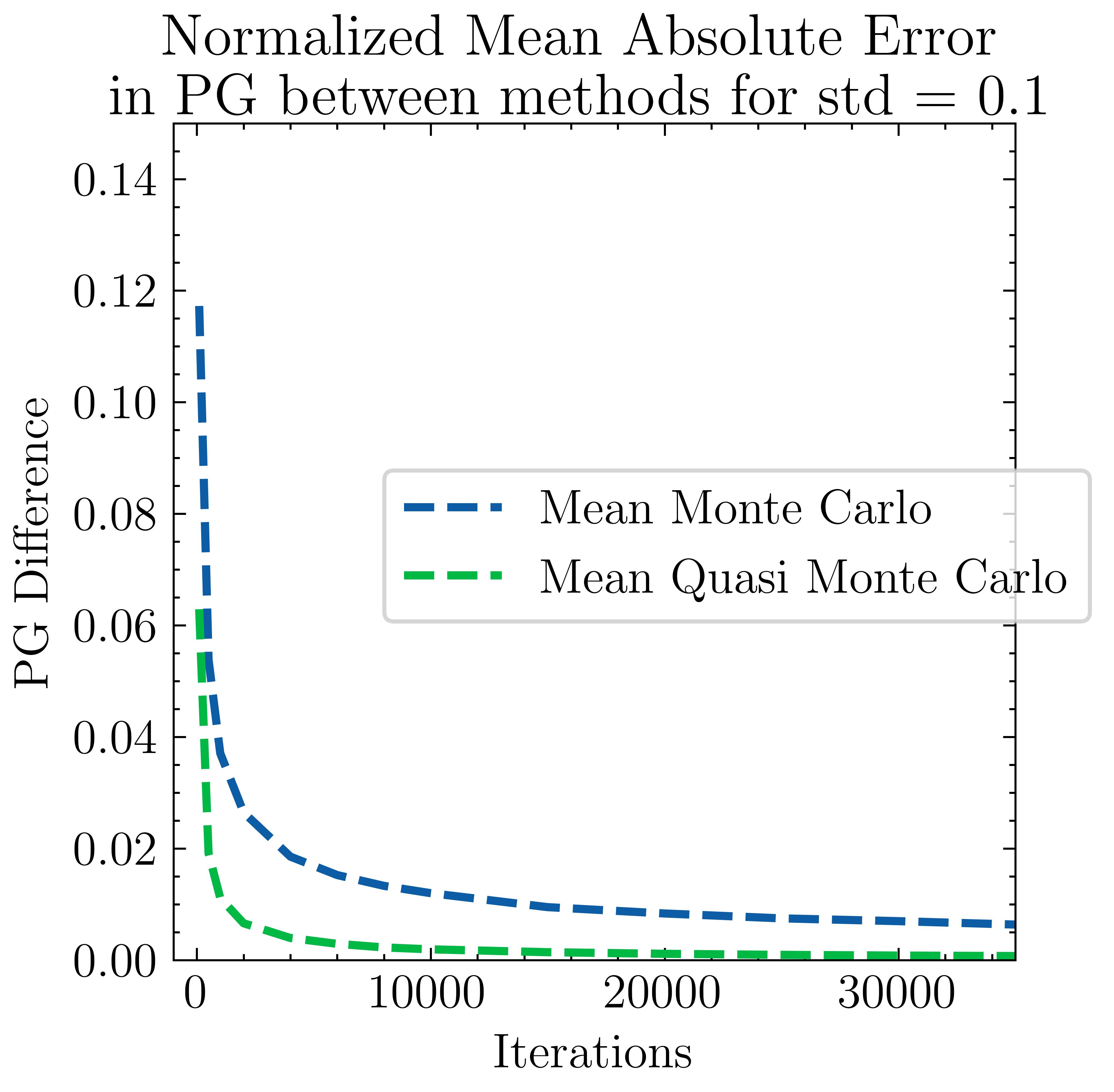}{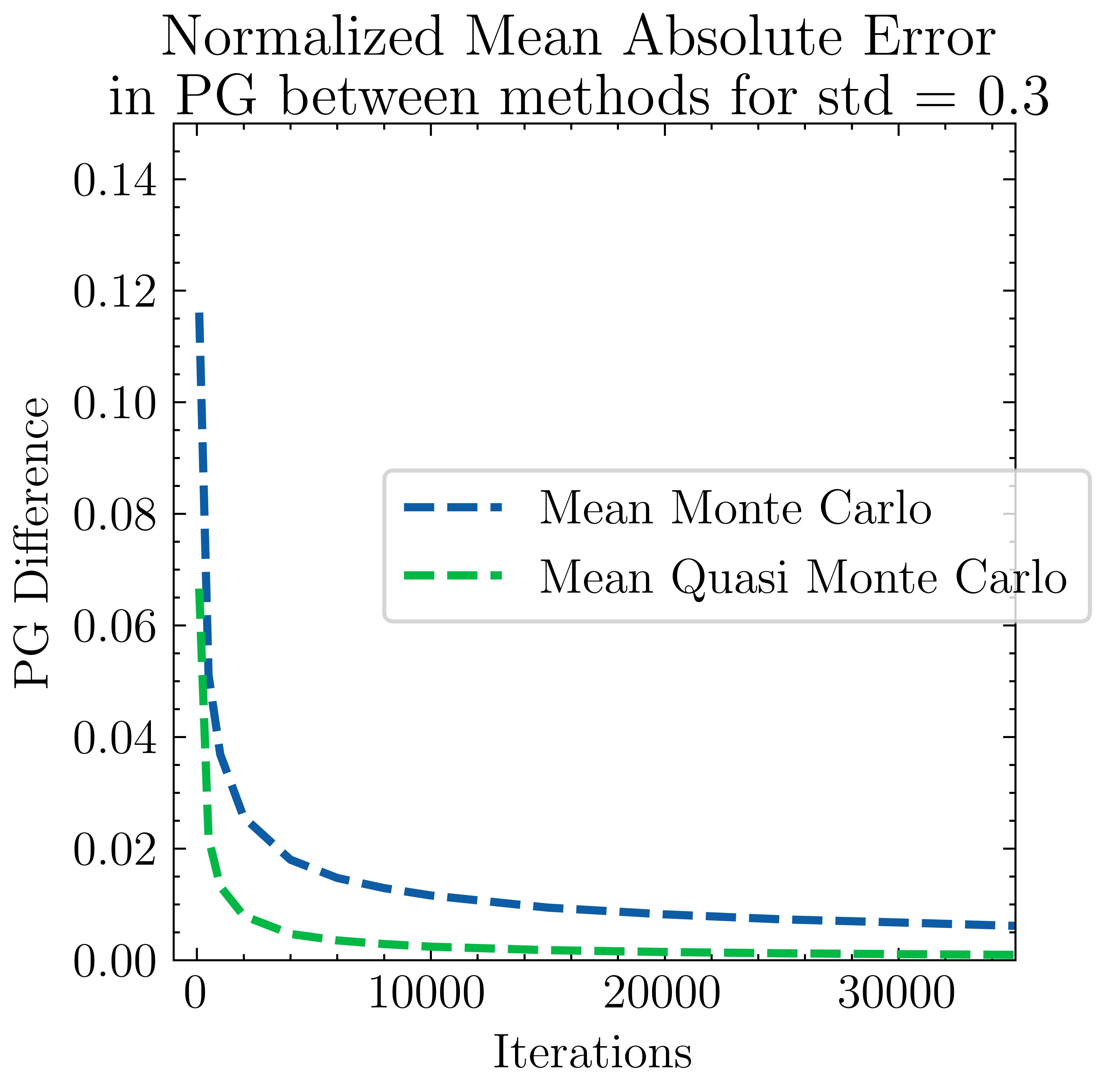}
    \includegraphics[width=0.48\textwidth]{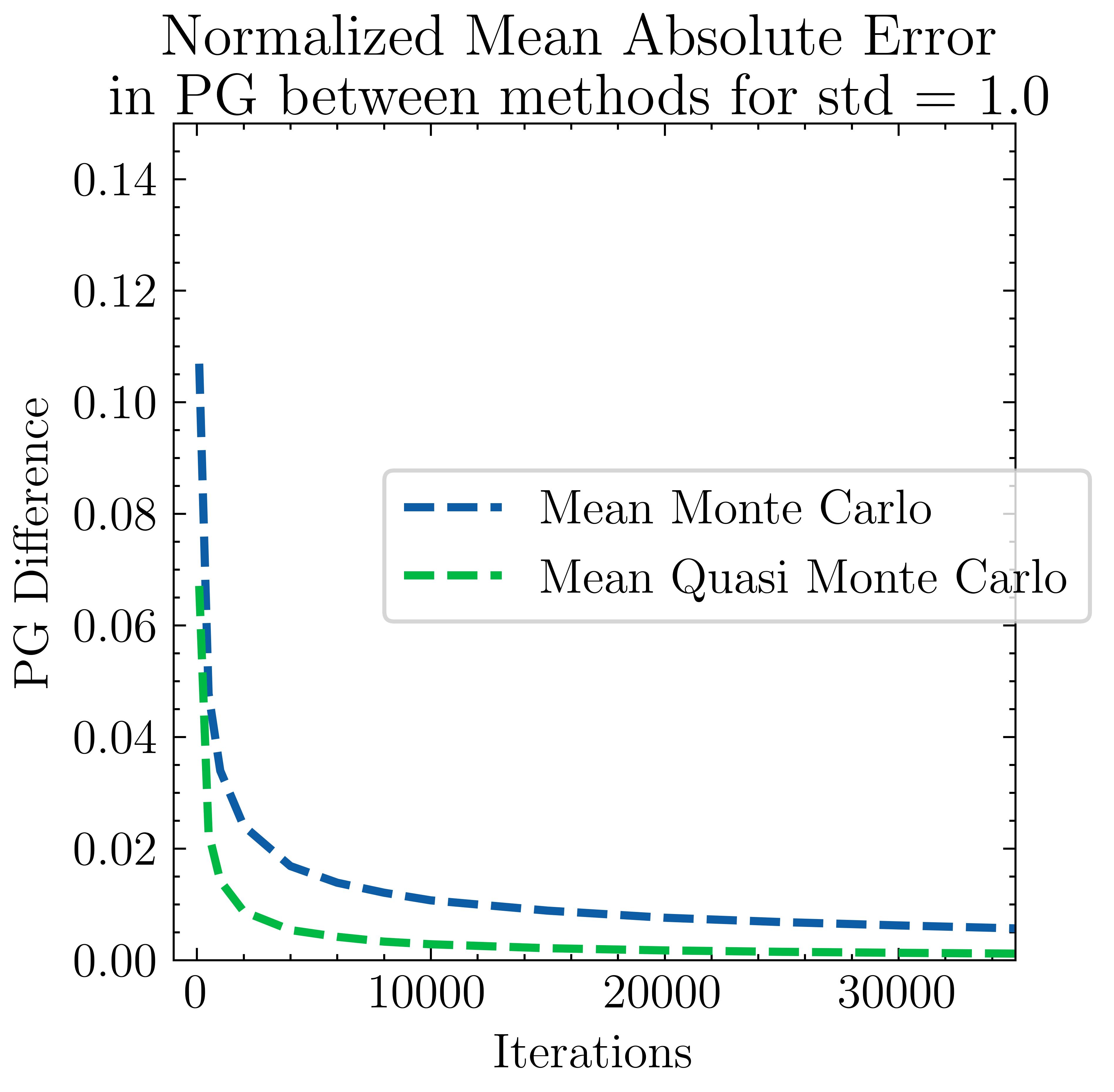}
    \caption{Bigger model for Parkinson Telemonitoring  dataset}
    \label{fig:}
\end{figure}
\clearpage
\subsection{Results from feature removing experiment}\label{a:noising_experiment}

% \begin{figure}
%     \centering
%     \includegraphics[width=1.0\linewidth]{figures/remove_wine_model.jpg}
%     \caption{Feature removing results for bigger Wine model}
%     \label{fig}
% \end{figure}
% Wine
% \begin{figure}
%     \centering
%     \includegraphics[width=1.0\linewidth]{figures/remove_housing_model.jpg}
%     \caption{Feature removing results for bigger Housing model}
%     \label{fig}
% \end{figure}

% \begin{figure}
%     \centering
%     \includegraphics[width=1.0\linewidth]{figures/remove_telemetry_model.jpg}
%     \caption{Feature removing results for bigger Parkinson Telemonitoring  model}
%     \label{fig}
% \end{figure}

% \subsection{Results from feature removing experiment}\label{a:noising_experiment}

% \begin{figure}
%     \centering
%     \includegraphics[width=1.0\linewidth]{figures/noising_wine_model.jpg}
%     \caption{Feature removing results for bigger Wine model}
%     \label{fig}
% \end{figure}
% Wine
% \begin{figure}
%     \centering
%     \includegraphics[width=1.0\linewidth]{figures/noising_housing_model.jpg}
%     \caption{Feature removing results for bigger Housing model}
%     \label{fig}
% \end{figure}

% \begin{figure}
%     \centering
%     \includegraphics[width=1.0\linewidth]{figures/noising_telemetry_model.jpg}
%     \caption{Feature removing results for bigger Parkinson Telemonitoring  model}
%     \label{fig}
% \end{figure}

\begin{table}[ht!]
    \centering
    \caption{Rankings comparison wrt. feature randomising metric}\label{tab:randomizing}
    \begin{tabular}{|c|c|c|c|c|}
      %\toprule  
      \bfseries  $k$ & SHAP & $\sigma$ = $0.1$ & $\sigma$ = $0.3$& $\sigma$ = $1.0$\\
      %\midrule  
\hline
      \multicolumn{5}{c}{Bigger  Red Wine Quality model} \\
            \hline
1 & 0.944 & 0.922 & 0.923 & 0.919 \\
2 & 0.893 & 0.902 & 0.881 & 0.891 \\
3 & 0.858 & 0.89 & 0.87 & 0.868 \\

\hline
      \multicolumn{5}{c}{Bigger  California Housing model} \\
            \hline
1 & 1.32e+05 & 1.412e+05 & 1.379e+05 & 1.264e+05 \\
2 & 1.3e+05 & 1.354e+05 & 1.321e+05 & 1.263e+05 \\
3 & 1.282e+05 & 1.318e+05 & 1.287e+05 & 1.27e+05 \\

\hline
      \multicolumn{5}{c}{Bigger  Parkinson Telemonitoring model} \\
            \hline
1 & 8.87 & 8.96 & 8.99 & 8.945 \\
2 & 8.748 & 8.882 & 8.882 & 8.906 \\
3 & 8.585 & 8.777 & 8.809 & 8.784 \\

      \hline
    \end{tabular}
\end{table}

\begin{table}[ht!]
    \centering
    \caption{Rankings comparison wrt. remove-and-retrain metric}\label{tab:retraining}
    \begin{tabular}{|c|c|c|c|c|}
      %\toprule  
      \bfseries  $k$ & SHAP & $\sigma$ = $0.1$ & $\sigma$ = $0.3$& $\sigma$ = $1.0$\\
      %\midrule  
\hline
      \multicolumn{5}{c}{Bigger PGI Red Wine Quality model} \\
            \hline
1 & 0.67 & 0.621 & 0.637 & 0.667 \\
2 & 0.72 & 0.638 & 0.66 & 0.672 \\
3 & 0.722 & 0.652 & 0.665 & 0.711 \\

\hline
      \multicolumn{5}{c}{Bigger PGI California Housing model} \\
            \hline
1 & 8.092e+04 & 7.187e+04 & 7.206e+04 & 7.485e+04 \\
2 & 8.913e+04 & 7.693e+04 & 8.062e+04 & 8.707e+04 \\
3 & 9.136e+04 & 8.267e+04 & 8.965e+04 & 9.665e+04 \\

\hline
      \multicolumn{5}{c}{Bigger PGI Parkinson Telemonitoring model} \\
            \hline
1 & 7.216 & 6.745 & 6.863 & 7.02 \\
2 & 7.388 & 6.951 & 7.138 & 7.292 \\
3 & 7.55 & 7.021 & 7.282 & 7.45 \\

      \hline
    \end{tabular}
\end{table}
\subsection{Results from PGI comparison experiment}\label{a:pgi_experiment}

\begin{table}[hb!]
    \centering
    \caption{Greedy $\PG^2$ vs SHAP rankings wrt. $\PGI^2$.}\label{tab:}
    \begin{tabular}{|c|c|c|c|c|}  
      \bfseries $\sigma'$ & SHAP & $\sigma$ = $0.1$ & $\sigma$ = $0.3$&  $\sigma$ = $1.0$\\
      \hline
      \multicolumn{5}{c}{Single Red Wine Quality model}\\
      \hline
      0.1 & 0.021 & 0.023 & 0.023 & 0.023 \\
0.3 & 0.077 & 0.095 & 0.096 & 0.096 \\
1.0 & 0.151 & 0.174 & 0.185 & 0.185 \\
\hline
\multicolumn{5}{c}{Single California Housing model}\\
\hline
0.1 & 3.267e+08 & 3.987e+08 & 3.987e+08 & 3.987e+08 \\
0.3 & 1.421e+09 & 1.582e+09 & 1.597e+09 & 1.597e+09 \\
1.0 & 5.921e+09 & 6.276e+09 & 6.451e+09 & 6.459e+09 \\
\hline
\multicolumn{5}{c}{Single Parkinson Telemonitoring model}\\
\hline
0.1 & 1.084 & 1.084 & 1.084 & 1.084 \\
0.3 & 3.438 & 3.573 & 3.823 & 3.823 \\
1.0 & 6.143 & 7.322 & 8.444 & 8.449 \\
\hline
    \end{tabular}
\end{table}
\end{document}